\documentclass{article}

\usepackage{arxiv}
\usepackage[utf8]{inputenc}
\usepackage{amsmath,amssymb,amsfonts}
\usepackage{graphicx}
\usepackage{hyperref}
\usepackage{cite}
\usepackage{booktabs}
\usepackage{amsthm}
\usepackage{multirow}
\usepackage{mathrsfs}
\usepackage{xcolor}
\usepackage{textcomp}
\usepackage{algorithm}
\usepackage{algorithmicx}
\usepackage{algpseudocode}
\usepackage{listings}
\usepackage{tikz}
\usetikzlibrary{positioning}
\usepackage{float}

\theoremstyle{plain}
\newtheorem{theorem}{Theorem}
\newtheorem{definition}{Definition}

\title{A Network Routing Approach to Supply Path Optimization in Programmatic Advertising}

\author{
    \textbf{Amelia B\u{a}dic\u{a}} \thanks{\texttt{amelia.badica@edu.ucv.ro}, Department of Statistics and Business Informatics, University of Craiova, Romania} \and 
    \textbf{Costin B\u{a}dic\u{a}} \thanks{\texttt{costin.badica@edu.ucv.ro}, Department of Computers and Information Technology, University of Craiova, Romania}
}
\date{\today}

\begin{document}

\maketitle

\begin{abstract}
Programmatic advertising relies on complex, multi-hop supply chains, creating an ``ad-tech tax'' that current Supply Path Optimization (SPO) heuristics struggle to resolve. We model this ecosystem as a Directed Acyclic Graph (DAG) using real-world OpenRTB data. To find the true network equilibrium, we introduce a two-speed architecture: an offline Mixed-Integer Linear Program (MILP) that approximates the logarithmic objective via SOS2 to establish a theoretical benchmark for Nash Social Welfare (NSW), alongside a fast K-Shortest Paths (KSP) heuristic for real-time execution. Testing on a programmatic DAG with 820 nodes and 2,854 arcs across a 50-publisher cohort reveals major flaws in existing methods. First, standard decentralized routing fails 74\%--76\% of the time due to cascaded fees and latency timeouts, halving typical welfare even when transactions clear. Second, simply minimizing hops (Shortest Path) is also inefficient: by ignoring buyer budgets, it caps transaction feasibility at 92\%. In contrast, our MILP approach optimizes clearing prices to reach 96\%--98\% feasibility with significantly higher NSW ($p < 10^{-13}$), proving that optimal paths require no more than 4 hops. Finally, sensitivity analysis confirms 100.0\% structural stability against reserve-price fluctuations; the optimal network absorbs financial shocks seamlessly through margin compression rather than path fragmentation.
\end{abstract}

\keywords{Supply Path Optimization \and Programmatic Advertising \and Directed Acyclic Graph \and Mixed-Integer Linear Programming \and Nash Social Welfare}

\section{Introduction} 

Programmatic advertising operates as a real-time, decentralized, multi-tier information system connecting publishers to advertisers through complex chains of Supply-Side Platforms (SSPs), ad exchanges, and programmatic resellers \cite{wang2017, choi2020}. When a user loads a webpage, an impression opportunity is auctioned within milliseconds via Real-Time Bidding (RTB) protocols. However, because each intermediary along the routing chain extracts a percentage-based fee, this multi-hop topology suffers from severe multi-tier marginalization. Empirical industry and academic audits demonstrate that up to 59\% of gross advertiser expenditure is absorbed by this cumulative ``ad-tech tax'' and unattributed fee leakage before reaching the content publisher \cite{ana2023, papadopoulos2018}. Addressing this systemic inefficiency requires moving beyond isolated bidding heuristics to model the supply chain as an integrated, multi-attribute network.

Within the information systems literature, the algorithmic complexity, transparency, and economic efficiency of programmatic advertising ecosystems have garnered significant attention. Recent studies have examined the mechanics of RTB strategies, valuation optimizers, multi-touch budget allocation across channels, and the design of automated market mechanisms for efficient price discovery \cite{balseiro2014, wang2017}. Despite these advances, state-of-the-art Supply Path Optimization (SPO) algorithms in both academic literature and commercial platforms rely almost entirely on isolated, single-agent optimization. As highlighted by \cite{choi2020}, existing models treat the broader intermediary network as an exogenous black box. They focus predominantly on local decisions, such as a Demand-Side Platform (DSP) optimizing bid shading in first-price auctions \cite{wang2017,balseiro2019learning}, an SSP dynamically adjusting reserve floor prices \cite{balseiro2014}, or localized competition under header bidding \cite{despotakis2020}, while ignoring the cross-routing topology of the intermediaries. Because buyers lack end-to-end visibility into intermediate take-rates, commercial SPO devolves into myopic, statistical heuristics (such as blanket-blocking paths with low historical win rates or arbitrarily capping hop counts) rather than solving for system-wide economic equilibrium.

Previous research in multi-agent systems and network economics formalized intermediation as Directed Acyclic Graphs (DAGs), proving that collectively profitable networks possess a unique pricing equilibrium maximizing Nash Social Welfare (NSW) \cite{badica2019jifs, badica2020cit, nash1950}. However, foundational intermediation models assume concurrent execution across all incoming supply arcs (\textit{AND} logic) and additive flat margins, which do not capture the operational realities of digital ad exchanges. This paper extends formal DAG intermediation theory to programmatic advertising by incorporating substitutable parallel routing (\textit{OR} logic induced by header bidding), multiplicative percentage-based fees, strict Service Level Agreement (SLA) latency timeouts, and a normalized logarithmic objective to ensure monetary scalability and fair surplus distribution across supply paths with a variable number of actors.

In live programmatic markets, routing occurs under severe information asymmetry, and clearing prices are driven by myopic, inter-buyer auction competition. While real-time auctions resolve transactions locally, uncoordinated multi-tier fee stacking frequently triggers severe marginalization that starves downstream intermediaries or violates publisher floor prices. To quantify the efficiency lost to this decentralized fragmentation, we propose a two-speed operational architecture: an offline Mixed-Integer Linear Program (MILP) that uses Special Ordered Sets of Type 2 (SOS2) to approximate the logarithmic objective, paired with a sub-second K-Shortest Paths (KSP) heuristic for real-time execution. By operating retrospectively on empirical OpenRTB logs where buyer valuations, reserve floors, and network latencies are already known, the centralized MILP computes the exact NSW-maximizing routing and clearing prices. This provides a concrete mathematical baseline to measure how much efficiency is lost by myopic, real-time heuristics.

To bridge the gap between theoretical mechanism design and empirical information systems research, we construct and evaluate our model using structurally realistic evaluation graphs derived from real-world OpenRTB \texttt{schain} and \texttt{sellers.json} datasets \cite{iab2022openrtb, vekaria2024}. Specifically, this paper makes five primary contributions:
\begin{itemize}
    \item \textbf{Normative Benchmark and Two-Speed Architecture for Multi-Tier SPO:} We formulate a novel MILP with logarithmic SOS2 approximation tailored to programmatic DAGs. This model synthesizes substitutable parallel routing (\textit{OR} logic), multiplicative percentage take-rates, and strict SLA latency bounds, and is paired with a sub-second KSP tactical heuristic for real-time path scoring.
    \item \textbf{Quantification of Systemic Inefficiency:} By benchmarking our centralized oracle against myopic decentralized routing strategies, we compute the Price of Anarchy (PoA), empirically demonstrating that uncoordinated routing fails in 74\%--76\% of transactions across all latency regimes while nearly halving typical geometric welfare even when transactions clear.
    \item \textbf{Structural Limits of Supply Chain Depth:} Through combinatorial state-space profiling, we establish $k=4$ hops as a practical upper bound for tractable programmatic routing. Extending routing depth beyond this threshold exponentially inflates computational complexity (exceeding $10^{17}$ theoretical paths), while our optimizations within this bound show that realized efficient paths naturally average only 2.13--2.22 hops.
    \item \textbf{Macro-Micro Structural Stability:} A two-tiered sensitivity analysis across 50 highly connected publishers demonstrates 100.0\% structural routing invariance against reserve-price perturbations, revealing that optimal topologies absorb financial shocks via internal margin compression in clearing prices rather than physical path fragmentation.
    \item \textbf{Suboptimality of Financial-Blind Shortest Paths:} Through computational experiments under heterogeneous fee structures, we demonstrate that purely structural heuristics that minimize hop counts ignore downstream buyer budgets, capping feasibility at 92\% and yielding significantly lower NSW compared to multi-hop routing with endogenous clearing-price optimization.
\end{itemize}

The remainder of this paper is organized as follows. Section \ref{sec:background} outlines the OpenRTB supply chain architecture and the limitations of current optimization approaches. Section \ref{sec:theory} formalizes the updated DAG intermediation model, integrating substitutable routing and proportional take-rates into an exact MINLP. Section \ref{sec:methodology} describes the algorithmic linearizations required for the offline solver and the online heuristic. Section \ref{sec:experiments} details the empirical data synthesis pipeline, evaluates the routing mechanisms against baseline heuristics, and discusses managerial implications. Finally, Section \ref{sec:conclusion} concludes the paper, summarizing key limitations and outlining directions for future research.

\section{Background and Motivation}\label{sec:background}

\subsection{The OpenRTB Supply Chain and Market Friction}
The programmatic advertising ecosystem operates as a distributed, real-time information system enforcing a one-way, sequential flow of ad inventory from publishers to advertisers, paired with a reverse flow of financial settlements. When a user visits a publisher's webpage, the browser or ad server generates an OpenRTB bid request containing user context, device telemetry, and placement specifications. This request is broadcast and routed through multiple layers of intermediaries (including SSPs, ad exchanges, and programmatic resellers) before reaching DSPs that bid on behalf of advertisers \cite{wang2017, choi2020}.

Historically, the opacity of these multi-tier transactions fostered severe market inefficiencies, including unauthorized inventory reselling, domain spoofing, and cascading intermediary fees that erode advertiser budgets \cite{ana2023, papadopoulos2018, bashir2019longitudinal}. To mitigate these vulnerabilities and establish auditability across distributed ad-tech infrastructures, the Interactive Advertising Bureau (IAB) introduced three mandatory supply-chain transparency standards \cite{iab2022openrtb}:

\begin{itemize}
    \item \textbf{Authorized Digital Sellers (\texttt{ads.txt}):} A publisher-hosted declarative ledger listing all SSPs and resellers authorized to monetize the publisher's inventory, explicitly distinguishing between direct contractual relationships (\texttt{DIRECT}) and third-party resellers (\texttt{RESELLER}).
    \item \textbf{Seller Transparency Registries (\texttt{sellers.json}):} A complementary JSON ledger hosted by SSPs and exchanges that maps internal account identifiers to legal corporate entities. Its \texttt{seller\_type} attribute directly maps operational business roles to our formal DAG partitions: origin content creators correspond to the \texttt{PUBLISHER} type (set $\mathcal{S}$), while SSPs, exchanges, and resellers correspond to the \texttt{INTERMEDIARY} or \texttt{BOTH} types (set $\mathcal{I}$).
    \item \textbf{SupplyChain Object (\texttt{schain}):} A dynamic JSON array embedded directly within live OpenRTB bid requests. Each intermediary that processes the request and participates in the financial settlement appends a structured node containing its Advertising System Identifier (\texttt{asi}) and Seller ID (\texttt{sid}), creating an immutable, hop-by-hop execution trace of the transaction path.
\end{itemize}

From an architectural standpoint, modern programmatic routing has largely transitioned from sequential ``waterfall'' setups, where ad requests were offered to SSPs one by one in a static priority order, to parallel \textit{header bidding} \cite{despotakis2020}. In header bidding (implemented either client-side via wrappers such as \texttt{Prebid.js} or server-to-server), publishers simultaneously solicit bids from multiple SSPs. Each SSP conducts its own internal auction across connected downstream partners and returns its highest clearing bid to the publisher. Crucially, while multiple parallel branches are evaluated concurrently, a single ad impression is indivisible; consequently, routing paths are mutually exclusive, and exactly one end-to-end chain executes the final transaction.

\begin{figure}[htbp]
\centering
{\begin{tikzpicture}[
    scale=0.72, transform shape,
    >=stealth,
    node distance=1.25cm and 1.7cm,
    entity/.style={rectangle, draw, thick, rounded corners, minimum width=1.8cm, minimum height=0.65cm, text centered, font=\sffamily},
    seller/.style={entity, fill=blue!10},
    intermediary/.style={entity, fill=orange!10},
    buyer/.style={entity, fill=green!10},
    arrow/.style={->, thick}
]
    \node[seller] (pub) {Publisher ($s$)};
    \node[intermediary, right=of pub, yshift=1.25cm] (ssp1) {SSP 1 ($I_1$)};
    \node[intermediary, right=of pub, yshift=-1.25cm] (ssp2) {SSP 2 ($I_2$)};
    \node[buyer, right=of ssp1, yshift=0.85cm] (dsp1) {DSP 1 ($b_1$)};
    \node[intermediary, right=of ssp1, yshift=-0.85cm] (adx) {Ad Exchange ($I_3$)};
    \node[intermediary, right=of ssp2] (reseller) {Reseller ($I_4$)};
    \node[buyer, right=of adx, yshift=-0.85cm] (dsp2) {DSP 2 ($b_2$)};
    \draw[arrow] (pub) to[out=0, in=180] (ssp1);
    \draw[arrow] (pub) to[out=0, in=180] (ssp2);
    \draw[arrow] (ssp1) to[out=0, in=180] (dsp1);
    \draw[arrow] (ssp1) to[out=0, in=180] (adx);
    \draw[arrow] (adx) to[out=0, in=180] (dsp2);
    \draw[arrow] (ssp2) to[out=0, in=180] (reseller);
    \draw[arrow] (reseller) to[out=0, in=180] (dsp2);
    \node at (current bounding box.north) [yshift=0.3cm] {};
\end{tikzpicture}}
\caption{A multi-tier OpenRTB supply chain mapped as a DAG. Arrows indicate the forward flow of the ad impression from the seller ($s$) through intermediaries ($I$) to the buyers ($b$), while financial payments and percentage fees propagate in reverse.\label{fig:schain_dag}}
\end{figure}
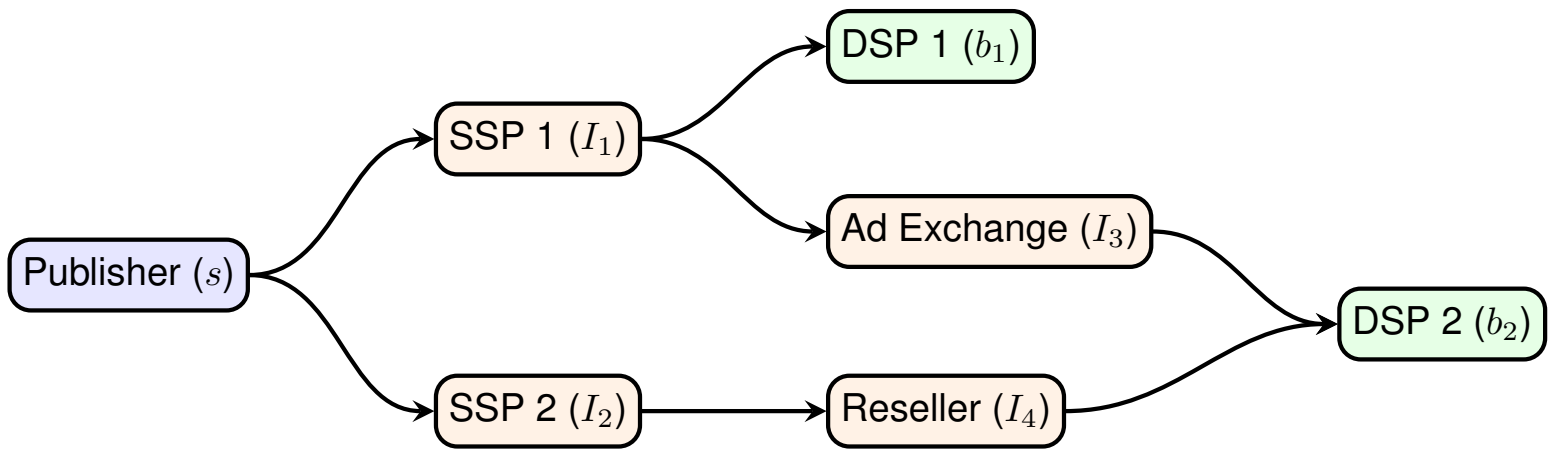

To illustrate this operational dynamic, Figure~\ref{fig:schain_dag} depicts a representative multi-tier header bidding topology. The Publisher ($s$) initiates a parallel auction by broadcasting the bid request simultaneously to two SSPs. SSP~1 ($I_1$) routes the opportunity both directly to DSP~1 ($b_1$) and indirectly through an Ad Exchange ($I_3$), whereas SSP~2 ($I_2$) routes the request via a Reseller ($I_4$) before reaching DSP~2 ($b_2$). This commercial architecture forms a DAG in which multiple routes compete for the same impression, yet only one complete supply chain (e.g., $\text{Publisher} \rightarrow \text{SSP 2} \rightarrow \text{Reseller} \rightarrow \text{DSP 2}$) clears the auction and is serialized into the winning \texttt{schain} payload.

Furthermore, this multi-tier routing is governed by strict temporal constraints. Because auctions must resolve before webpage rendering completes, publishers enforce rigid SLA timeouts (typically 200--300~ms in browser environments, or slightly wider windows in server-side setups). Every intermediary hop introduces network propagation delay and local auction processing overhead. Consequently, an economically attractive buyer positioned at the end of a deep reseller chain may be rendered infeasible if cumulative transit latencies breach the publisher's global timeout threshold. Together, \texttt{ads.txt}, \texttt{sellers.json}, and \texttt{schain} eliminate the need for purely synthetic topologies, providing an empirical, verifiable blueprint of commercial interconnections and real-time latency telemetry for applied DAG optimization.

\subsection{Limitations of Current Approaches}
Despite the rapid evolution of programmatic infrastructure, existing academic research in operations research, computer science, and information systems addresses digital advertising primarily through isolated, single-agent optimization models or purely descriptive empirical audits. We categorize the limitations of the current literature along four distinct dimensions:

\subsubsection{Single-Agent Bidding and Yield Optimization}
The vast majority of computational advertising literature focuses on optimizing the local strategy of an individual market participant while treating the surrounding supply chain as an exogenous black box \cite{choi2020}:
\begin{itemize}
    \item \textit{Buyer-Side Pacing and Bid Shading:} Demand-side models concentrate on \textit{budget pacing} (dynamically throttling bid rates to distribute finite campaign funds smoothly over time \cite{balseiro2019learning}) and \textit{bid shading} (where DSPs use statistical estimation in first-price auctions to bid below their true valuation without sacrificing win probability \cite{wang2017,balseiro2019learning}).
    \item \textit{Seller-Side Yield Management:} Supply-side formulations model the exchange's problem as dynamic reserve price optimization, setting localized floor prices to maximize short-term revenue \cite{balseiro2014}. Similarly, game-theoretic analyses of header bidding \cite{despotakis2020} examine how parallel auctions intensify competition and alter bidding strategies, yet they abstract away the multi-hop intermediary network behind a single auction layer.
\end{itemize}

Within the information systems domain, studies have also investigated automated price discovery mechanisms in continuous double auctions \cite{balseiro2014} and strategies to bypass suboptimal RTB performance optimizers \cite{wang2017}. However, because these models assume direct or single-layer intermediation, they fail to capture the multiplicative margin erosion (\textit{double marginalization}) that occurs when multiple intermediaries sequentially deduct percentage take-rates along a multi-hop path.

\subsubsection{Heuristic SPO in Industry}
To mitigate fee stacking and duplicate bid requests, DSPs and media agencies have adopted commercial SPO practices \cite{choi2020, ana2023}. However, intermediaries rarely disclose their exact percentage take-rates ($\gamma_i$) within standard OpenRTB bid requests, creating severe information asymmetry \cite{papadopoulos2018, iab2022openrtb}. Combined with the NP-hard combinatorial complexity of evaluating multi-attribute graphs within millisecond auction windows, this lack of transparency forces industry SPO to rely on myopic, brute-force heuristics. Common practices include capping maximum hop counts (such as blocking paths with more than two intermediaries), pruning supply paths based solely on historical win rates, or defaulting to topologically shortest paths \cite{vekaria2024, choi2020}. As our empirical results demonstrate, purely structural heuristics are financially blind: minimizing hop counts frequently routes traffic through high-fee premium exchanges, paradoxically inflating the total ad-tech tax.

\subsubsection{Descriptive Transparency Audits vs. Normative Equilibrium}
A growing stream of empirical measurement research leverages IAB transparency files (\texttt{ads.txt} and \texttt{sellers.json}) to map the macroscopic topology of the ad-tech ecosystem, uncovering dark inventory pooling, unauthorized reselling, and systemic fee leakage \cite{vekaria2024, papadopoulos2018, bashir2019longitudinal}. While these studies provide vital empirical visibility into \textit{potential} commercial relationships (the static availability graph) and historical anomalies, they remain strictly descriptive. They lack the concurrent integration of real-time clearing prices, buyer budgets, publisher reserve floors, and hop-by-hop latency bounds required to compute a formal economic routing equilibrium.

\subsubsection{Classical Intermediation Networks vs. Programmatic Realities}
Finally, existing network intermediation models in multi-agent systems \cite{badica2019jifs, badica2020cit} establish the theoretical foundation for NSW maximization on DAGs, but rely on deterministic \textit{AND}-node execution and additive margin structures. They do not account for mutually exclusive path selection (\textit{OR}-node header bidding), multiplicative fee cascades, or hard end-to-end latency constraints.

In summary, the literature lacks a normative, network-wide optimization framework capable of jointly evaluating topological routing, multiplicative fee cascades, and physical latency limits. By extending formal DAG intermediation theory with substitutable OR-routing and SLA constraints, our MILP formulation bridges this gap, establishing a rigorous theoretical benchmark for multi-tier SPO.

\section{Theoretical Framework: Updating the DAG Model}\label{sec:theory}

To address the complexities of SPO in programmatic advertising, we must expand the baseline intermediation framework. We first formalize the foundational DAG model, then introduce the mathematical adjustments required to capture header bidding logic and proportional fee structures.

\subsection{The Baseline Intermediation Model}
An intermediation network can be formalized as a DAG, denoted as $G=\langle\mathcal{V},\mathcal{A}\rangle$ \cite{badica2019jifs, badica2020cit}. The set of nodes is partitioned into three disjoint sets: $\mathcal{V} = \mathcal{S} \cup \mathcal{I} \cup \mathcal{B}$, representing sellers, intermediaries, and buyers, respectively. The set of directed arcs $\mathcal{A} \subseteq (\mathcal{S} \cup \mathcal{I}) \times (\mathcal{I} \cup \mathcal{B})$ represents the network topology, where the direction of each arc corresponds to the forward flow of goods (ad inventory), while financial payments flow in the reverse direction from buyers to sellers.

Each arc $e \in \mathcal{A}$ is mapped to a continuous price variable $p_e \in \mathbb{R}^+$. For any node $v \in \mathcal{V}$, $IN(v)$ and $OUT(v)$ are the sets of incoming and outgoing arcs. In the baseline model, routing follows an \textit{AND} logic: to produce an output, an intermediary must acquire inputs from all connected incoming arcs.\looseness=-1 

The utility $u_i$ of an intermediary $i \in \mathcal{I}$ is the absolute margin between revenues and costs:
\begin{equation}
    u_i = \sum_{e \in OUT(i)} p_e - \sum_{e \in IN(i)} p_e
\end{equation}

Similarly, the utilities for sellers $s \in \mathcal{S}$ and buyers $b \in \mathcal{B}$ are defined relative to their respective limit prices. A seller's utility is the revenue generated minus their minimum acceptable floor price ($\sigma_s$), and a buyer's utility is their maximum valuation or budget ($\beta_b$) minus the actual purchase price:\looseness=-1
\begin{equation}
\begin{aligned}
    u_s &= \sum_{e \in OUT(s)} p_e - \sigma_s \\
    u_b &= \beta_b - \sum_{e \in IN(b)} p_e
\end{aligned}
\end{equation}

The network is \textit{collectively profitable} if a price vector yields non-negative utility ($u_v \ge 0$ for all $v \in \mathcal{V}$), bounding prices between seller reserves ($\sigma$) and buyer budgets ($\beta$). Because the feasible utility space forms a convex polytope and the logarithmic objective ($\sum \log u_v$) is concave, a unique price vector exists that maximizes the system's NSW \cite{badica2020cit}.

\subsection{Header Bidding and Substitutable Routing (OR Nodes)}
The baseline model assumes that all incoming arcs to a node are executed concurrently (\textit{AND} logic). In traditional supply chains, this represents assembling multiple components or aggregating product bundles when economically viable. However, in programmatic advertising, a single ad impression is indivisible. When a publisher or an SSP runs a header bidding auction, they receive bids from multiple parallel paths and they must select the single highest bid.\looseness=-1

Because the ad impression cannot be split, this exclusivity applies to both sides of the transaction. A specific seller $s \in \mathcal{S}$ initiates the auction by generating a single indivisible ad impression, and the network must route it to exactly one winning buyer from the available set $\mathcal{B}$. 

To model this, we introduce two sets of binary decision variables. For each arc $e \in \mathcal{A}$, let $z_e \in \{0, 1\}$ be the discrete decision to route the transaction through that specific path. Additionally, for each buyer $b \in \mathcal{B}$, let $x_b \in \{0, 1\}$ be the selection of buyer $b$ as the unique winner of the auction, subject to the exclusive selection constraint $\sum_{b \in \mathcal{B}} x_b = 1$. To ensure the active paths form a single, unbroken chain from the initiating seller $s$ to the chosen buyer, inventory generated at $s$ must flow through intermediate nodes without loss and be absorbed exclusively by the selected buyer where $x_b = 1$.\looseness=-1

\begin{itemize}
    \item \textbf{Decision Variables:} The model determines the active structural routing paths ($z_e$), the unique buyer selection ($x_b$), and the continuous transaction prices along the paths ($p_e$).
    \item \textbf{Constraints:} The constraints enforce the DAG topology, the unit flow conservation of the indivisible impression, the sellers' minimum reserve prices ($\sigma$), and the buyers' maximum budgets ($\beta$).
    \item \textbf{Objective:} We maximize NSW as $\max \sum_{v \in \mathcal{V}} w_v$, where $w_v = \log(u_v / \beta_{\max})$ for active nodes (with $\beta_{\max} = \max_{b \in \mathcal{B}} \beta_b$) and $w_v = 0$ otherwise.
\end{itemize}

Optimizing for this normalized NSW serves two goals: selecting the most efficient path to maximize total surplus and calculating prices to distribute that surplus fairly. The normalization guarantees monetary scale invariance \cite{nash1950} (the routing equilibrium remains independent of the chosen currency magnitude) and proportional fairness \cite{kelly1998rate}. Assuming the selected supply path is financially viable (i.e., the selected buyer's budget, discounted by the take-rates of all active intermediaries, strictly exceeds the seller's reserve), the logarithmic function acts as an interior-point barrier. This ensures that the utility $u_v$ for every active node $v$ remains strictly positive ($u_v > 0$). Because $u_v \le \beta_{\max}$, the evaluated logarithmic weight $w_v$ is inherently non-positive ($w_v \le 0$), acting as a structural penalty for each additional active hop, while unselected nodes safely contribute zero. Finally, because this objective multiplies binary routing variables with continuous prices, the problem naturally formulates as a Mixed-Integer Nonlinear Program (MINLP).

\subsection{Percentage Take-Rates and Non-Linear Utilities}
Consistent with standard ad-tech monetization models, intermediaries apply a percentage-based take-rate (ad-tech tax) rather than flat fees \cite{ana2023, balseiro2019learning}. Let $\gamma_i \in (0, 1)$ be intermediary $i$'s fee. Their utility is proportional to the gross revenue from the buyer side (outgoing arcs):\looseness=-1
\begin{equation}
    u_i = \gamma_i \sum_{e \in OUT(i)} z_e p_e
\end{equation}

They pass the remaining payment fraction to the seller side, enforcing financial flow conservation:
\begin{equation}
    \sum_{e \in IN(i)} z_e p_e = (1 - \gamma_i) \sum_{e \in OUT(i)} z_e p_e
\end{equation}

This structure drives multi-tier marginalization: across a path $I_1, \dots, I_n$, the price reaching the seller is multiplicatively reduced by $(1 - \gamma_1)\dots(1 - \gamma_n)$. 

Programmatic auctions execute in real-time during webpage loads, subjecting routing decisions to tight physical constraints, primarily latency. Publishers and ad exchanges impose hard timeout limits (typically 200--300 milliseconds) to prevent ad loading delays from degrading the user experience.\looseness=-1

We expand arc variables to vectors $(p_e, t_e)$, where $t_e \in \mathbb{R}^+$ is the hop's continuous network delay. Header bidding enforces a hard continuous threshold $T_{\max,s} \in \mathbb{R}^+$ set by the initiating seller $s$, discarding paths exceeding this limit regardless of profitability:
\begin{equation}
    \sum_{e \in \mathcal{A}} z_e t_e \le T_{\max,s}
\end{equation}

\subsection{Complete MINLP Formulation}

This section formally defines the SPO problem as an MINLP.
\begin{definition}\label{def:spo}
We model the SPO problem for a transaction initiated by seller $s \in \mathcal{S}$ (with reserve $\sigma_s$ and latency SLA $T_{\max,s}$) across the intermediation DAG $G$, competing to select exactly one winning buyer $b \in \mathcal{B}$, where each buyer $b \in \mathcal{B}$ has budget $\beta_b$. Let $\beta_{\max} = \max_{b \in \mathcal{B}} \beta_b$ denote the maximum budget available in the network. We optimize routing $(z_e)$, buyer selection ($x_b$) and prices $(p_e)$ to maximize NSW subject to structural flow \eqref{eq:struct_flow}, single-buyer selection \eqref{eq:buyer_select}, financial balance \eqref{eq:fin_flow}, latency limits \eqref{eq:latency_threshold}, utilities \eqref{eq:util_i}--\eqref{eq:util_b}, the logarithmic objective mapping \eqref{eq:util_w}, and variable domains \eqref{eq:domains}:
\begin{equation}
    \max \sum_{v \in \mathcal{V}} w_v \label{eq:objective}
\end{equation}
\begin{subequations}
\begin{align}
    \llap{\text{s.t.} \quad} & \sum_{e \in OUT(v)} z_e - \sum_{e \in IN(v)} z_e = 
    \begin{cases} 
      1 & \text{if } v = s \\
     -x_v & \text{if } v \in \mathcal{B} \\
      0 & \text{otherwise} 
    \end{cases} \label{eq:struct_flow} \\
    & \sum_{b \in \mathcal{B}} x_b = 1, \quad x_b \in \{0,1\} \label{eq:buyer_select} \\
    & \sum_{e \in IN(i)} z_e p_e = (1 - \gamma_i) \sum_{e \in OUT(i)} z_e p_e 
    \quad \forall i \in \mathcal{I} \label{eq:fin_flow} \\
    & \sum_{e \in \mathcal{A}} z_e t_e \le T_{\max, s} \label{eq:latency_threshold} \\
    & u_i = \gamma_i \sum_{e \in OUT(i)} z_e p_e 
    \quad \forall i \in \mathcal{I} \label{eq:util_i} \\
    & u_{s'} = \left(\sum_{e \in OUT(s')} z_e p_e\right) - \sigma_{s'} \left(\sum_{e \in OUT(s')} z_e\right) 
    \,\,\,\, \forall s' \in \mathcal{S} \label{eq:util_s} \\
    & u_b = \beta_b x_b - \left(\sum_{e \in IN(b)} z_e p_e\right) 
    \quad \forall b \in \mathcal{B} \label{eq:util_b} \\
    & w_v = 
    \begin{cases}
      \log(u_v / \beta_{\max}) & u_v > 0 \\
      0 & u_v = 0
    \end{cases}
    \quad \forall v \in \mathcal{V} \label{eq:util_w} \\
    & z_e \in \{0, 1\}, \quad 0 \le p_e \le \beta_{\max} \quad \forall e \in \mathcal{A} \nonumber \\
    & u_v \ge 0 \quad \forall v \in \mathcal{V} \label{eq:domains}
\end{align}
\end{subequations}
\end{definition}

Unit flow conservation ensures an unbroken path from sellers to buyers. The piecewise mapping of $w_v$ keeps the objective defined over all graph nodes by assigning zero weight to inactive nodes. By normalizing the utility $u_v$ against the global maximum budget $\beta_{\max}$ in Equation \eqref{eq:util_w}, the fraction $u_v / \beta_{\max}$ is bounded in $(0, 1]$, rendering the logarithmic terms strictly non-positive ($w_v \le 0$). This provides two properties: it guarantees scale invariance regardless of the absolute monetary units used (e.g., dollars versus cents), and it introduces a structural penalty. Because every active node adds a negative term to the objective sum, the solver only extends a supply path if it reaches a buyer with a sufficiently larger budget to offset the penalty of the additional hop. While the clearing price on a singular pre-determined path could be computed in closed form, our mechanism optimizes structural routing ($z_e$) and continuous prices ($p_e$) simultaneously. This joint decision inherently creates the non-linear product ($z_e p_e$). Consequently, solving for the true network equilibrium strictly requires this MINLP formulation and its subsequent Big-M and SOS2 approximations.

\subsection{Existence of Feasible Solutions}
Before analyzing the computational difficulty of the SPO problem, we must formally define the conditions under which a valid transaction can occur. In programmatic advertising, an auction fails to clear if the buyer's budget cannot cover both the intermediaries' fees and the publisher's reserve price. We formulate this economic reality as a necessary and sufficient feasibility condition.
\begin{theorem}\label{thm:feasibility}
    The SPO MINLP (Definition \ref{def:spo}) admits at least one feasible solution with a well-defined, finite logarithmic NSW objective (i.e., strictly positive utility $u_v > 0$ for every active participant) if and only if there exists at least one directed path $P$ in $\mathcal{A}$ from the initiating seller $s$ to some buyer $b \in \mathcal{B}$ that satisfies both:
    \begin{enumerate}
        \item \textbf{Latency Compliance:} $\sum_{e \in P} t_e \le T_{\max, s}$
        \item \textbf{Financial Viability:} $\beta_b \prod_{i \in P \setminus \{s, b\}} (1 - \gamma_i) > \sigma_s$
    \end{enumerate}
\end{theorem}

\begin{proof}
    We prove this by analyzing the utility constraints of the active path. 
    
    ($\Rightarrow$ Necessity): Assume a feasible solution exists with a finite logarithmic NSW objective $\sum_{v \in P} \log(u_v / \beta_{\max}) > -\infty$. By definition of the logarithmic function, every active node $v \in P$ must receive a strictly positive utility ($u_v > 0$). Furthermore, the flow conservation constraints ensure that active nodes $P$ form a path from $s$ to exactly one selected buyer $b$ (where $x_b = 1$). The latency condition $\sum_{e \in P} t_e \le T_{\max, s}$ follows directly from the model constraints. 
    
    For financial viability, let $p_{b}$ be the price paid by the selected buyer $b$ to the final intermediary on path $P$ (or directly to $s$ if $P$ has a single hop). Since $u_b = \beta_b - p_{b} > 0$, we know $p_{b} < \beta_b$. As this payment travels upstream to the seller, the financial flow constraints dictate that each intermediary $i \in P \setminus \{s, b\}$ passes $(1 - \gamma_i)$ of the received payment to the next upstream node. Thus, the final price reaching the seller is $p_{s} = p_{b} \prod_{i \in P \setminus \{s, b\}} (1 - \gamma_i)$. Since the seller's utility must be strictly positive ($u_s = p_{s} - \sigma_s > 0$), we have $p_{s} > \sigma_s$. Substituting $p_{b} < \beta_b$ yields $\beta_b \prod_{i \in P \setminus \{s, b\}} (1 - \gamma_i) > p_{b} \prod_{i \in P \setminus \{s, b\}} (1 - \gamma_i) = p_{s} > \sigma_s$. Thus, $\beta_b \prod_{i \in P \setminus \{s, b\}} (1 - \gamma_i) > \sigma_s$ must hold.

    ($\Leftarrow$ Sufficiency): Assume there exists a path $P$ satisfying both conditions, terminating at a specific buyer $b \in \mathcal{B}$, with intermediary take-rates $\gamma_i \in (0, 1)$. We construct a feasible solution by setting $z_e = 1$ for all $e \in P$ and $x_b = 1$ for this terminal buyer. For all other edges and buyers, we set $z_e = 0$ and $x_{b'} = 0$. This satisfies the structural flow, exclusive buyer selection, and latency constraints. Because $\beta_b \prod_{i \in P \setminus \{s, b\}} (1 - \gamma_i) > \sigma_s$, there exists a strictly positive total surplus along the path. We can select a continuous buyer payment $p_b$ such that $\sigma_s \left( \prod_{i \in P \setminus \{s, b\}} (1 - \gamma_i) \right)^{-1} < p_b < \beta_b$. Propagating this price upstream according to the financial flow constraints yields $p_s > \sigma_s$, $0 < p_b < \beta_b$, and strictly positive flow through every intermediary. Hence, $0 < u_v < \beta_{\max}$ for every active node $v \in P$, ensuring a well-defined, finite logarithmic NSW objective ($w_v = \log(u_v / \beta_{\max}) \in (-\infty, 0)$) while unselected nodes ($x_{b'} = 0$) contribute $w_{b'} = 0$. 
\end{proof}

This theorem guarantees that solver infeasibility is not a formulation failure, but rather indicates that no mutually beneficial trade exists under the given market conditions.

\subsection{Computational Complexity}

Combining latency constraints with financial routing logic renders the problem computationally intractable for large-scale multi-tier networks. 
\begin{theorem}\label{thm:complexity}
    The SPO problem (Definition \ref{def:spo}) is NP-hard.
\end{theorem}

\begin{proof}
    We prove this by reduction from the decision version of the Weight-Constrained Shortest Path Problem restricted to DAGs (WCSPP-DAG), which remains NP-complete~\cite{garey1979computers, ahuja1993network}. We must show that any arbitrary instance of WCSPP-DAG can be transformed into an instance of our SPO problem in polynomial time.

    Let an instance of the WCSPP-DAG decision problem be defined by a DAG $G_0 = (V_0, E_0)$, a source node $s$, a target node $b$, strictly positive integer costs $c_e \ge 1$ and non-negative weights $w_e$ for each edge $e \in E_0$, a maximum cost limit $C$, and a maximum weight limit $W$. The problem asks if there exists a path from $s$ to $b$ such that the total weight is at most $W$ and the total cost is at most $C$.

    We construct an SPO instance as follows. Map the vertices of $G_0$ directly into our network, where $s$ is the initiating seller and $b$ is the sole available buyer ($\mathcal{B} = \{b\}$, forcing $x_b = 1$). Because take-rates in the SPO model are vertex attributes of intermediaries whereas costs in WCSPP-DAG are edge attributes, we subdivide each directed edge $e = (u, v) \in E_0$ by inserting a dedicated intermediary node $i_e$ with two directed arcs $(u, i_e)$ and $(i_e, v)$. This transformation preserves the DAG structure and adds $|E_0|$ nodes and $2|E_0|$ arcs in polynomial time. Set arc latencies $t_{(u, i_e)} = w_e$ and $t_{(i_e, v)} = 0$, and set the seller's latency SLA to $T_{\max,s} = W$. This enforces the weight constraint $\sum_{(u,v) \in P} w_{(u,v)} \le W$ via the latency threshold constraint.

    To map the additive cost constraint $\sum_{e \in P} c_e \le C$ to the multiplicative financial flow, set the buyer budget $\beta_b = 1$ (so $\beta_{\max} = \beta_b = 1$), set the seller reserve $\sigma_s = e^{-(C + 0.5)}$, and assign each intermediary node $i_e$ a take-rate $\gamma_{i_e} = 1 - e^{-c_e} \in (0, 1)$, so that $-\log(1 - \gamma_{i_e}) = c_e$.

    By Theorem~\ref{thm:feasibility}, a path $P$ admits continuous prices yielding strictly positive active utilities ($0 < u_v < 1$, and thus a finite normalized logarithmic NSW objective $\sum_{v \in P} \log(u_v / \beta_{\max}) = \sum_{v \in P} \log(u_v) > -\infty$) if and only if $\sum_{e \in P} c_e \le C$ is satisfied via the financial viability condition:
    \begin{align*}
    \beta_b \prod_{e \in P} (1 - \gamma_{i_e}) > \sigma_s 
    &\iff \prod_{e \in P} e^{-c_e} > e^{-(C + 0.5)} \\
    &\iff e^{-\sum_{e \in P} c_e} > e^{-(C + 0.5)}.
    \end{align*}
    
    Taking the negative natural logarithm yields $\sum_{e \in P} c_e < C + 0.5$. Because edge costs $c_e$ are integers, any path cost $\sum_{e \in P} c_e$ is an integer, so $\sum_{e \in P} c_e < C + 0.5$ is equivalent to $\sum_{e \in P} c_e \le C$.

    Therefore, a feasible SPO routing path yielding a valid, finite logarithmic NSW objective exists if and only if there is a "yes" answer to the WCSPP-DAG decision problem (i.e., a path satisfying both $C$ and $W$ exists). Because merely finding a feasible solution with a finite objective is NP-hard, the broader SPO optimization problem (maximizing that objective) is NP-hard. Since the reduction runs in $O(|V_0| + |E_0|)$ time, the proof is complete.
\end{proof}

\subsection{Illustrative Numerical Example}
To ground the general MINLP formulation in practical OpenRTB data, we instantiate the mathematical model for the specific DAG presented in Figure \ref{fig:schain_dag}. 

The parameters are derived from standard programmatic data sources: the seller's minimum reserve ($\sigma_s$) is set via the publisher's \texttt{bidfloor}, the buyers' maximum budgets ($\beta_b$) represent the DSPs' bid responses, the take-rates ($\gamma_i$) reflect industry-standard intermediary fees, and the latency values ($t_e$) represent network telemetry measured in milliseconds.

Let the maximum latency for this seller be $T_{\max, s} = 150$~ms. The individual parameters for the network participants are: 
\begin{itemize}
\item \textbf{Seller cutoff price:} $\sigma_s = 0.50$; 
\item \textbf{Intermediaries' fee rates:} $\gamma_{I_1} = 0.10, \gamma_{I_2} = 0.15, \gamma_{I_3} = 0.20, \gamma_{I_4} = 0.10$; 
\item \textbf{Buyer budgets:} $\beta_1 = 1.50, \beta_2 = 2.00$, establishing the global normalizing bound $\beta_{\max} = 2.00$; 
\item \textbf{Latencies ($t_e$):} Each discrete network hop incurs a 40~ms latency, except the reseller connections which incur 60~ms.
\end{itemize}

The unit flow and selection constraints enforce mutually exclusive routing from the seller and a unique buyer choice:
\begin{align}
    z_{s,I_1} + z_{s,I_2} &= 1 \\
    x_{b_1} + x_{b_2} &= 1
\end{align}

To illustrate the multiplicative financial flow, consider the three-hop path routing through SSP 2 and the Reseller to DSP 2. This margin degradation is modeled by linking prices from buyer to seller:\looseness=-1
\begin{align}
    p_{I_4, b_2} &\le \beta_2 \cdot x_{b_2} && \text{(Buyer constraint at } b_2\text{)} \\
    p_{I_2, I_4} &= (1 - 0.10) \cdot p_{I_4, b_2} && \text{(Financial flow at } I_4\text{)} \\
    p_{s, I_2} &= (1 - 0.15) \cdot p_{I_2, I_4} && \text{(Financial flow at } I_2\text{)} \\
    p_{s, I_2} &\ge \sigma_s \cdot z_{s, I_2} && \text{(Seller constraint at } s\text{)}
\end{align}

If this path is activated and DSP 2 is selected as the exclusive winner ($z_{s,I_2} = z_{I_2,I_4} = z_{I_4,b_2} = 1$ and $x_{b_2} = 1$, which forces $x_{b_1} = 0$), the financial constraints force the price reaching the seller to be $p_{s, I_2} = (0.85)(0.90) \cdot p_{I_4, b_2} = 0.765 \cdot p_{I_4, b_2}$. The total latency of this path is $40 + 60 + 40 = 140$~ms, which satisfies $T_{\max,s} \le 150$. Because $0.765 \beta_2 = 1.53 > \sigma_s = 0.50$, Theorem \ref{thm:feasibility} holds.

The solver concurrently evaluates this 3-hop path to $b_2$ against the more direct 2-hop path to DSP 1 ($s \to I_1 \to b_1$). The direct path to $b_1$ incurs only a single 10\% take-rate, preserving 90\% of the budget, but connects to a buyer with a lower maximum valuation ($\beta_1 = 1.50$, yielding a maximum feasible seller price of $1.35$). In contrast, the 3-hop path accesses a premium budget ($\beta_2 = 2.00$, yielding $1.53$ to the seller) but forces the system to distribute the resulting surplus among four active nodes instead of three, while operating near the latency limit (140~ms vs. 80~ms). 

By maximizing the normalized objective $\sum \log(u_v / \beta_{\max})$, the MINLP resolves this trade-off. Because the scale-invariant logarithmic barrier severely penalizes distributing surplus across unnecessary intermediaries, the solver determines that the $0.18$ CPM difference in net seller value is insufficient to fairly compensate the additional Reseller node ($I_4$). Consequently, the optimal solution sets $x_{b_1} = 1$ and $x_{b_2} = 0$, activating the routing variables $z_{s,I_1} = z_{I_1,b_1} = 1$. This corresponds to the winning path $s \to I_1 \to b_1$. The system successfully prunes the expensive multi-tier path, selecting DSP 1 as the unique winner to maximize the NSW of the shorter, more efficient chain.

\section{Computational Methodology}\label{sec:methodology}
Because the SPO MINLP is NP-hard, solving it on large-scale networks requires reformulating the non-linearities. To utilize standard commercial solvers, we linearize the bilinear terms and approximate the logarithmic objective via SOS2, transforming the model into a MILP.

\subsection{Linearization of Bilinear Routing Variables}
The model's non-convexity arises from bilinear terms in the utility constraints, where binary routing decisions ($z_e$) multiply continuous prices ($p_e$). To enable standard linear solvers, we linearize this by introducing an auxiliary continuous variable $y_e = z_e p_e$.

We apply the standard Big-M linearization technique \cite{nemhauser1988integer} to resolve the bilinear terms. Let $M$ be a large enough upper bound; for our domain, the maximum budget across all buyers, $\beta_{\max} = \max_{b \in \mathcal{B}} \beta_b$, serves as Big-M parameter. The bilinear product $z_e p_e$ is thus replaced by an auxiliary continuous variable $y_e$, constrained by the following system of linear inequalities:\looseness=-1
\begin{subequations}\label{eq:big_m}
\begin{align}
    y_e &\le \beta_{\max} z_e \label{eq:bm_upper_z} \\
    y_e &\le p_e \label{eq:bm_upper_p} \\
    y_e &\ge p_e - \beta_{\max} (1 - z_e) \label{eq:bm_lower} \\
    y_e &\ge 0 \label{eq:bm_zero}
\end{align}
\end{subequations}

When the path is inactive ($z_e = 0$), inequalities (\ref{eq:bm_upper_z}) and (\ref{eq:bm_zero}) force $y_e = 0$, ensuring no financial transfer occurs. When the path is active ($z_e = 1$), inequalities (\ref{eq:bm_upper_p}) and (\ref{eq:bm_lower}) force $y_e = p_e$. To complete the MILP transformation, we explicitly substitute $y_e$ for every instance of $z_e p_e$ in the financial flow and utility equations (Eqs. \ref{eq:fin_flow}, \ref{eq:util_i}, \ref{eq:util_s}, and \ref{eq:util_b}). This exactly replicates the original MINLP logic while ensuring all constraints remain strictly linear for the solver.

\subsection{Piecewise Linear Approximation of the Objective}\label{sec:piecewise_linear}
The second computational challenge is the logarithmic objective function. Because the Big-M linearization already forces $y_e = 0$ (and consequently $u_v = 0$) for inactive nodes, we can approximate the logarithm over the active domain using SOS2 \cite{beale1970special, williams2013model}. This replaces the non-linear curve with piecewise linear segments, allowing standard MILP solvers to compute the objective while preserving its concavity.

First, we introduce a node-activation variable $x_v \in \{0, 1\}$ linked to the routing variables:
\begin{equation}\label{eq:node_activation}
\begin{aligned}
    x_v &= \sum_{e \in OUT(v)} z_e && \forall v \in \mathcal{S} \cup \mathcal{I} \\
    x_b &= \sum_{e \in IN(b)} z_e && \forall b \in \mathcal{B}
\end{aligned}
\end{equation}

To ensure the solver can evaluate the poorest valid paths without arbitrary truncation, we first define a minimum strictly positive utility threshold $U_1$. Assuming there exists at least one buyer with a strictly positive budget ($\{b \in \mathcal{B} \mid \beta_b > 0\} \neq \emptyset$), we set $U_1 = \epsilon \cdot \min_{\{b \in \mathcal{B} \mid \beta_b > 0\}} \beta_b$, where $\epsilon \in (0, 1)$ is a small tolerance scalar that ensures $0 < U_1 < \beta_{\max}$. We then define a discrete set of breakpoints $k \in \{1, \dots, K\}$ that are geometrically spaced across the feasible utility range $[U_1, \beta_{\max}]$:
\begin{equation}\label{eq:log_grid}
    U_k = U_1 \left(\frac{\beta_{\max}}{U_1}\right)^{\frac{k-1}{K-1}} \quad \forall k \in \{1, \dots, K\}
\end{equation}

Because the logarithmic objective exhibits steep curvature near zero, a uniform arithmetic grid would suffer from severe interpolation error on low-utility segments, allowing the solver to underestimate the penalty of small participant margins. By spacing breakpoints geometrically, consecutive intervals maintain a constant logarithmic step $\Delta \log U = \frac{1}{K-1} \log(\beta_{\max}/U_1)$, ensuring uniform approximation accuracy across all orders of magnitude. 

The utility $u_v$ and its normalized logarithmic value $w_v$ are represented via SOS2 variables $\lambda_{v,k}$:
\begin{subequations}\label{eq:sos2}
\begin{align}
    u_v &= \sum_{k=1}^K \lambda_{v,k} U_k && \forall v \in \mathcal{V} \\
    w_v &= \sum_{k=1}^K \lambda_{v,k} \log(U_k / \beta_{\max}) && \forall v \in \mathcal{V} \\
    \sum_{k=1}^K \lambda_{v,k} &= x_v && \forall v \in \mathcal{V} \\
    u_v &\ge U_1 \cdot x_v && \forall v \in \mathcal{V} \\
    \lambda_{v,k} &\ge 0 && \forall v \in \mathcal{V},\; k \in \{1, \dots, K\}
\end{align}
\end{subequations}

The sum of weights equals $x_v$. By definition of SOS2, at most two adjacent variables $\lambda_{v,k}$ can be non-zero. When a node is active ($x_v = 1$), weights form a proper convex combination, and utility is lower-bounded by $U_1$. When inactive ($x_v = 0$), the constraints force all $\lambda_{v,k} = 0$, yielding $w_v = 0$, safely removing unselected nodes from the NSW objective without triggering a mathematical error.

This piecewise approximation introduces a slight, bounded deviation from the true continuous optimal solution. The magnitude of this error is controlled by the density of the breakpoint grid ($K$). While a higher $K$ drives the approximation error asymptotically to zero, it proportionally increases the solver's branching complexity, requiring a practical trade-off between exactness and computational speed. We empirically evaluate this trade-off in Section \ref{sec:experiments} to determine a practical baseline for $K$ that balances execution time with objective accuracy.

\section{Experimental Evaluation}\label{sec:experiments}
To validate the MINLP formulation, we conduct experiments using a hybrid evaluation graph: empirical OpenRTB topologies augmented with semi-synthetic constraints. This section details the data collection for verifiable DAGs, explains the algorithmic synthesis of missing financial parameters, outlines the solver configuration, and compares optimal NSW routing against standard heuristics.\looseness=-1

\subsection{Data Acquisition and DAG Construction}\label{sec:data_acquisition}

Rather than relying on purely synthetic network topologies, we construct our DAG and parameter data set directly from empirically grounded programmatic data. To evaluate the proposed MINLP formulation, the solver requires four specific data elements for each routing scenario, which must be mapped to their corresponding graph components:

\begin{itemize}
    \item \textbf{Structural topology (\texttt{schain}):} The valid DAG pathways from the publishers to the buyers.\looseness=-1
    \item \textbf{Terminal financial limits ($\sigma, \beta$):} The seller's minimum reserve ($\sigma$) and the buyers' maximum budgets ($\beta$), assigned exclusively to the input and output nodes of the DAG.
    \item \textbf{Network latency ($t_e$):} Assigned to the edges, representing the temporal delay incurred per discrete network hop.
    \item \textbf{Intermediary take-rates ($\gamma$):} The fractional fees retained by transit nodes (SSPs, Ad Exchanges, and Resellers).
\end{itemize}

Because exact intermediary take-rates are proprietary and unobserved in bid responses, the $\gamma$ parameters are synthetically assigned within plausible market ranges and evaluated under both homogeneous (uniform) and heterogeneous (node-specific) fee settings. However, the structural topology, buyer budgets ($\beta$), and edge latencies ($t_e$) are empirically obtained, while the publisher reserve prices ($\sigma$) are semi-synthetically derived. This hybrid approach ensures that the model's structural and temporal constraints reflect real-world auction dynamics, while allowing for standardized financial testing.\looseness=-1

Publicly available RTB datasets \cite{wang2017, choi2020} and registry-based supply chain analyses \cite{vekaria2024} do not concurrently capture real-time multi-hop routing topologies, unmasked financial limits, and hop-by-hop latencies. To determine the most viable approach for obtaining these concurrent variables, we evaluated three data acquisition methods:

\begin{enumerate}
    \item \textbf{Public RTB Datasets (Academic/Open-Source):} Benchmark repositories frequently utilized in computational advertising literature (e.g., the Criteo or iPinYou datasets \cite{wang2017}) offer massive historical logs. However, they are fundamentally incompatible with multi-tier SPO. First, because these datasets were collected circa 2013–2014, long before the IAB \texttt{schain} standard became mandatory in 2020, we cannot reconstruct the exact multi-hop DAG topology. Second, to protect commercial interests, they routinely normalize bid prices to a $[0,1]$ scale or obscure them entirely, destroying the absolute monetary bounds ($\sigma, \beta$) needed for financial flow conservation. Finally, they omit network-level latency telemetry ($t_e$).
    \item \textbf{Commercial Data Aggregators:} Industry services (e.g., Sellers.guide, Jounce Media) map supply paths by cross-referencing public \texttt{ads.txt} and \texttt{sellers.json} ledgers. While these tools successfully map \textit{potential} business partnerships (the maximum available graph), they do not log \textit{executed} transactions. Consequently, they provide static, theoretical routes entirely devoid of the real-time auction payloads, i.e. the discrete bid prices ($\sigma, \beta$) and actual hop-by-hop latencies ($t_e$), that are required to compute our objective function.
    \item \textbf{Client-Side Interception (Browser Automation):} Deploying a headless browser to visit publisher websites and intercept the JSON objects handled by the header bidding wrapper (e.g., \texttt{Prebid.js}) provides native OpenRTB requests and responses, containing the complete \texttt{schain} object for winning routes, raw prices ($\sigma, \beta$), and precise latencies ($t_e$). However, scaling this approach is computationally expensive, and active bot-detection systems frequently block automated traffic. Furthermore, many publishers utilize non-standard wrapper implementations or obfuscate the global \texttt{Prebid} object, preventing reliable extraction of the auction data. Additionally, the extraction process is highly fragile, as runtime errors in the publishers' client-side scripts or minor updates to their site structure frequently break the automation.
\end{enumerate}

\subsubsection{The Data Synthesis Pipeline}

Based on this evaluation, the only viable method to obtain the precise structural and numerical parameters concurrently is to combine the second and third approaches: utilizing client-side interception anchored by server-side registry validation. To ensure academic rigor, handle the massive scale of the web, and prevent combinatorial explosion, our custom computational methodology (including data collection) is structured into a five-phase pipeline:

\paragraph{Phase 1: Seed Aggregation and Heuristic Filtering.} To ensure a statistically significant and unbiased starting point, we aggregate the top domains from three distinct global traffic rankings (Tranco, Majestic, and Cisco Umbrella). The merged dataset yielded a total of 2,411,502 unique domains. This raw list was then offline-filtered to exclude non-commercial top-level domains (e.g., \texttt{.gov}, \texttt{.edu}) and explicitly block spam or adult keywords, establishing a clean, commercial baseline of 2,203,723 domains ready for active probing.

\paragraph{Phase 2: Protocol Validation.} To efficiently process the aggregated domains, we deploy an asynchronous HTTP crawler configured with 400 concurrent workers and a matching TCP connection pool. First, utilizing 5-second timeouts, the crawler requests the \texttt{ads.txt} file for each domain, keeping only those that return an HTTP 200 status and contain explicit \texttt{DIRECT} or \texttt{RESELLER} string declarations. This step reduces the candidate pool to 143,965 domains. Second, the concurrent workers fetch the homepage for this remaining subset, scanning only the first 150 KB of the HTML payload to detect the active presence of client-side header bidding wrappers (specifically searching for \texttt{prebid}, \texttt{pbjs}, \texttt{owpbjs}, \texttt{vpbjs}, or \texttt{kargoPbjs} keywords). This file-size truncation prevents memory bottlenecks across the parallel connections and ensures rapid execution. Because header bidding scripts must execute early in the page lifecycle to resolve auctions before content rendering, their initialization is typically located within the document's \texttt{<head>}, well inside the 150 KB threshold. The risk of this heuristic is a minor false-negative rate for publishers with bloated HTML structures or footer-loaded wrappers; however, discarding these structural anomalies is acceptable to establish a standardized and strictly validated baseline of 5,315 publishers ready for active interception.\looseness=-1

\paragraph{Phase 3: Client-Side Telemetry Interception.} We deploy an asynchronous Playwright crawler to autonomously navigate the 5,315 validated publishers. To balance execution speed against the severe memory constraints of full DOM rendering, the crawler operates with a concurrency limit of 5 parallel workers. Rather than simple HTTP threads, each worker instantiates an entirely isolated Chromium browser context, enabling simultaneous navigation and JavaScript execution without state overlap or cache interference. 

To bypass bot-detection systems and ensure the header bidding wrappers trigger natural auction behavior, we deploy several evasion techniques. The crawler masks the automated environment by overriding the \texttt{navigator.webdriver} property, defining standard User-Agent profiles, injecting realistic geolocation data, and simulating user engagement via automated cookie-consent acceptance and viewport scrolling. To prevent memory bottlenecks across the parallel workers, network traffic is filtered by intercepting and aborting all heavy resource requests (images, media, and fonts).\looseness=-1

For the actual data extraction, the script scans the main document and any nested \texttt{iframes} to detect active header bidding wrappers using known global aliases (\texttt{pbjs}, \texttt{owpbjs}, \texttt{vpbjs}, \texttt{kargoPbjs}). Once an active object is located, the crawler injects a promise into the wrapper's command queue (\texttt{window[alias].que.push}), effectively hooking into the auction lifecycle and forcing the browser to wait for the \texttt{auctionEnd} event. The comprehensive auction payload is then extracted natively using the \texttt{getBidResponses()} method. 

The \texttt{originalBidfloor} is omitted from client-side auctions. Although a reserve price ($\sigma > 0$) exists to cover costs, publishers hide browser-side floors to prevent ``bid shading,'' shifting dynamic pricing server-side. Thus, empirical reserve prices appear null and must be synthetically estimated in Phase 5 to provide valid solver constraints.\looseness=-1

From the resulting JSON arrays, we map the empirical data directly to our MINLP parameters: the structural routing topology (via the \texttt{schain.nodes} array), the buyer's clearing Cost Per Mille (CPM) ($\beta$ via \texttt{cpm}), and the hop-by-hop network latency ($t_e$ via \texttt{timeToRespond}). 

This client-side methodology carries inherent operational risks. The reliance on DOM execution introduces vulnerability to page load timeouts (capped at 30 seconds), runtime JavaScript errors on the publisher's end, and highly customized wrapper implementations that successfully hide the bid responses. Furthermore, missing or malformed \texttt{schain} strings require fallback heuristics, such as mapping dead-end intermediaries to known buyers or defaulting to direct links, to enforce graph consistency. Despite these challenges, the extracted data is serialized into two strict schemas to feed the optimization solver. The dataset of graph nodes records the \texttt{node\_id}, \texttt{node\_type} (Publisher, Intermediary, or Buyer), \texttt{system\_domain}, and \texttt{name}. The transaction dataset logs the \texttt{auction\_id}, \texttt{publisher\_id}, the ordered \texttt{schain\_path}, \texttt{buyer\_id}, the buyer's clearing CPM ($\beta$ via \texttt{cpm}), and the measured latency ($t_e$ via \texttt{timeToRespond}). To construct the baseline topology, we aggregated these transaction logs, consolidating duplicate paths by summing their transaction volumes and averaging their latency telemetry. This processing yielded a validated empirical DAG comprising exactly 506 unique nodes and 594 unique directed arcs. Structural analysis of this baseline graph revealed a maximum path depth of exactly 2 hops (Publisher $\to$ Intermediary $\to$ Buyer), confirming that modern browser-level header bidding inherently flattens the supply chain.\looseness-1

\paragraph{Phase 4: Server-Side Topology Expansion.} The empirical data from Phase 3 provides the necessary buyer budgets ($\beta$) and baseline latency parameters ($t_e$). However, the inherent 2-hop limit of client-side interception results in an incomplete and disconnected DAG, which is insufficient to evaluate the MINLP solver against a realistic OpenRTB supply chain. To reconstruct a realistic end-to-end routing DAG, we expand the intermediate routing layer by aggregating the official IAB transparency registries into a structured server-side dataset.

To ensure the registry data is relevant to our empirical baseline, we executed a dedicated data acquisition pipeline prior to graph construction:
\begin{enumerate}
    \item \textbf{Publisher Registry Interrogation:} By parsing the \texttt{ads.txt} endpoints of the 5,315 validated publishers from Phase 2, we extracted over 5 million \texttt{DIRECT} and \texttt{RESELLER} declarations.
    \item \textbf{Empirical Intersection:} Filtering this massive registry exclusively to the 390 active publisher nodes captured in our Phase 3 transactions retained 589,373 declarations, mapping to 1,400 unique authorized ad-tech domains.
    \item \textbf{Server-Side Topology Extraction:} Finally, querying the \texttt{sellers.json} endpoints of these 1,400 ad-tech domains yielded 134,557 verified seller accounts (retaining only those designated as \texttt{INTERMEDIARY} or \texttt{BOTH}).
\end{enumerate}

This data acquisition pipeline produced the consolidated dataset (\texttt{sellers\_json\_network.csv}) containing the verified upstream-to-downstream mappings. During the algorithmic pipeline execution, this dataset is parsed to construct the raw server-side routing graph $H = (\mathcal{V}_H, \mathcal{A}_H)$. A directed edge is added from the listed selling entity (\texttt{domain}) to the hosting exchange or SSP (\texttt{ad\_tech\_domain}), provided the strings represent distinct commercial entities. Deduplicating these pairs yields a raw $H$ topology comprising 26,221 unique routing nodes and 63,106 declared edges. In Phase 5, this structure is merged with the empirical client-side graph $G$.\looseness=-1

\paragraph{Phase 5: Algorithmic Synthesis of the Semi-Synthetic Evaluation Graph.} 
This phase marks the entry point of our reproducibility pipeline. By merging the shallow client-side data extracted in Phase 3 with the extended server-side topology reconstructed in Phase 4, we build a complete end-to-end DAG. To ensure sufficient combinatorial complexity for the solver, we employ an Operations Research methodology: generating a semi-synthetic evaluation graph anchored in this combined empirical data \cite{rardin2001experimental}.\looseness-1

Formally, the algorithm receives two input sets. The first is the baseline empirical DAG extracted in Phase 3, denoted as $G = (\mathcal{V}_G, \mathcal{A}_G)$, which provides the validated terminal nodes representing the publishers, and the buyers with their clearing budgets ($\beta$). The second is the raw server-side topology extracted in Phase 4, denoted as $H = (\mathcal{V}_H, \mathcal{A}_H)$, containing the expansive, multi-tier intermediary network. The objective is to fuse these inputs into a final, computationally tractable evaluation graph $F = (\mathcal{V}_F, \mathcal{A}_F)$ to serve as the foundational input instance for the subsequent SPO MINLP experiments.

Because transparency registries ($H$) map supply-side relationships and rarely define terminal DSPs (buyers), a direct union of $G$ and $H$ would produce a fragmented network with dead-end paths. To resolve this, we deterministically connect these dead-end intermediaries to buyers when merging the two graphs. Specifically, each empirical buyer $b$ is parameterized with its maximum observed clearing bid ($\beta_b = \max \text{cpm}_b$) to represent its upper budget ceiling. We then evaluate the out-degree of all intermediary nodes in the unified graph $F$. Any intermediary node with an out-degree of zero is programmatically linked to exactly one empirical terminal buyer from $G$ (sorted lexicographically). The target buyer is selected preferentially via substring matching between the intermediary's domain and the buyer's identifier; if no direct match exists, a fallback buyer is drawn uniformly at random from the lexicographically sorted set $\mathcal{B}_G$ under a fixed pseudo-random seed (\texttt{seed=42} in our experiments) to ensure deterministic reproducibility. Each server-side edge from $H$ and each newly created bridging edge is assigned the network's mean empirical latency ($\bar{t}_e$). This ensures every intermediary has a valid terminal exit, completing the supply paths and guaranteeing end-to-end routing feasibility.

Once fused and bridged, the resulting larger network $F$ must be topologically pruned to guarantee computational feasibility for the solver and to satisfy real-world latency constraints. Naive cycle detection and path enumeration on networks exceeding 100,000 nodes trigger severe combinatorial explosions. Therefore, we implement a pruning pipeline parameterized by a topological depth limit ($k$) and a baseline viability latency threshold ($T_{\max}$):
\begin{enumerate}
    \item \textbf{DAG Enforcement:} We utilize a single-pass Depth-First Search (DFS) on the newly fused network $F$, operating in $O(|\mathcal{V}_F| + |\mathcal{A}_F|)$ time, to identify and remove back-edges. This step converts $F$ into a DAG, eliminating bid loops to align with our underlying network model.
    \item \textbf{Neighborhood Depth Bounding ($k$):} To reflect the shallow reachability of programmatic supply chains while preserving cross-tier routing options, we apply a bounded Breadth-First Search (BFS) of depth $k$ originating from the publisher nodes. This retains only the induced subgraph of nodes reachable within $k$ hops of at least one publisher (alongside the terminal buyers), while still permitting longer multi-hop cross-routing paths among retained intermediaries within the latency budget.\looseness=-1
    \item \textbf{Latency SLA Pruning ($T_{\max}$):} Using bidirectional shortest-path distances (Dijkstra's algorithm from publishers and reverse Dijkstra to buyers), we remove any edge $(u, v)$ whose minimum achievable end-to-end latency exceeds the scenario's uniform SLA budget $T_{\max}$ (where $T_{\max,s} = T_{\max}$ across all publishers $s \in \mathcal{S}$ in Equation~\ref{eq:latency_threshold}). Because this bound is exact, it shrinks the search space while preserving every feasible routing path.
    \item \textbf{Dead-End Pruning:} We recursively remove any intermediary nodes that were left with an out-degree of zero after the previous pruning steps, ensuring all remaining paths in the final evaluation graph successfully reach a buyer.
\end{enumerate}

Since exact publisher reserve prices ($\sigma$) are unobservable in the client-side data, they must be estimated to provide realistic financial constraints for the MINLP solver. The reserve price ($\sigma$) for each publisher is drawn from a truncated normal distribution based on the clearing budgets ($\beta$) of all buyers across the entire network. The distribution is strictly bounded between the 25th ($Q_1$) and 50th ($Q_2$) percentiles of the global $\beta$ values, with the mean set to $\mu = \frac{Q_1 + Q_2}{2}$ and the standard deviation to $\frac{Q_2 - Q_1}{4}$. We use this global market distribution to reflect practical publisher behavior: baseline reserve prices are anchored to macro-market liquidity rather than restricted to currently active supply paths. This also prevents publishers with limited connections in our dataset from receiving skewed reserve prices. \textit{To ensure that our structural conclusions are not biased by this synthetic generation, we rigorously test the robustness of the $\sigma$ parameter in a dedicated sensitivity analysis.}

Finally, to ensure valid solver inputs and enforce identical evaluation rules across all tested algorithms, any missing or non-positive empirical values ($t_e \le 0$ or $\beta_b \le 0$) are treated as missing data and deterministically replaced with the empirical arithmetic mean computed strictly over the valid, strictly positive observations ($\bar{t}_e$ for $t_e > 0$ and $\bar{\beta}$ for $\beta_b > 0$). All parameters are saved and evaluated in their raw empirical units (CPM for financial bounds and milliseconds for edge latencies), because the normalization logic embedded directly in the MILP objective \eqref{eq:util_w} natively resolves scale invariance without altering the structural inputs.

\subsubsection{Benchmark Instances and Scenario Selection}
To configure the benchmarking scenarios, we executed a preliminary parameter sweep over the pruning pipeline. Profiling the BFS reachability cutoff across $k \in \{3, 4, 5, 6\}$ revealed that expanding the induced neighborhood beyond 4 hops from the publishers exponentially inflates the combinatorial state space without yielding viable efficiency gains, aligning with the operational reality that legitimate OpenRTB intermediaries sit within a shallow hop radius of the publisher. Consequently, we fixed the neighborhood depth cutoff at $k=4$ for all evaluation graphs.

For the uniform latency SLA budget ($T_{\max, s} = T_{\max}$), we profiled the network across an interval of $[1500, 5000]$~ms. From this distribution, we selected three distinct configurations to serve as our experimental scenarios: $T_{\max} = 3000$~ms (representing a highly constrained structural network), $T_{\max} = 3500$~ms (a standard programmatic network), and $T_{\max} = 3800$~ms (the absolute combinatorial limit of our empirical dataset, beyond which further network expansion becomes marginal). 

Applying this configured pipeline onto the raw initial datasets (which comprised 26,221 nodes and 63,106 connections) yielded three distinct DAG instances. The structural and combinatorial characteristics for each of these three evaluation graphs ($F_{3000}$, $F_{3500}$, and $F_{3800}$) are compared in Table \ref{tab:graph_metrics}. Although synthetic scale-free generation models (such as Barabási-Albert preferential attachment) were not explicitly enforced, the empirical union of \texttt{ads.txt} and \texttt{sellers.json} topologies naturally exhibits heavy-tailed, power-law degree distributions characterized by a structural core of highly connected hub intermediaries alongside a long tail of fringe nodes. Notably, while the BFS tightly limits the shortest path from the publisher, cross-routing between intermediaries creates an extreme combinatorial explosion, scaling from approximately one thousand feasible paths in the constrained scenario to over 113 quadrillion in the relaxed scenario. These verified instances serve as the direct inputs for the SPO MINLP solver.

\begin{table}[htbp]
\centering
\caption{Topological and Combinatorial Characteristics of the Evaluation Graphs at $k=4$\label{tab:graph_metrics}}
{\begin{tabular}{lccc}
\hline
\textbf{Metric} & \textbf{$F_{3000}$} & \textbf{$F_{3500}$} & \textbf{$F_{3800}$} \\ \hline
\multicolumn{4}{c}{ \textit{Structural Properties}} \\ \hline
Total Nodes ($|\mathcal{V}_F|$) & 552 & 689 & 820 \\
Total Directed Arcs ($|\mathcal{A}_F|$) & 754 & 1,513 & 2,854 \\
Publisher Nodes (Sources) & 359 & 376 & 384 \\
Buyer Nodes (Sinks) & 58 & 58 & 58 \\
Intermediary Nodes & 135 & 255 & 378 \\
Maximum In-Degree & 65 & 70 & 86 \\
Maximum Out-Degree & 14 & 34 & 55 \\
Network Density & 0.0025 & 0.0032 & 0.0042 \\
Average Shortest Path Length & 2.58 & 3.50 & 3.48 \\ \hline
\multicolumn{4}{c}{ \textit{Combinatorial Complexity}} \\ \hline
Total Feasible Supply Paths & 1,044 & 8,346,364 & 113,235,630,309,802,258 \\
Global Path Length (Min/Max hops) & 2 / 5 & 2 / 22 & 2 / 89 \\ \hline
\multicolumn{4}{c}{ \textit{Financial \& Performance Parameters}} \\ \hline
Synthesized Reserve Price $\sigma$ (Avg CPM) & 0.0945 & 0.0947 & 0.0946 \\
Empirical Clearing Budget $\beta$ (Avg CPM) & \multicolumn{3}{c}{11.8405} \\
Routing Edge Latency $t_e$ (Avg ms) & 543.03 & 593.03 & 614.35 \\ \hline
\end{tabular}}
\end{table}

Although the MINLP model is formulated as a single-source optimization problem, routing inventory from one specific seller (publisher) to multiple buyers, the empirical evaluation graphs ($F_{3000}$, $F_{3500}$, $F_{3800}$) encapsulate global ecosystems with hundreds of origin nodes. To evaluate ecosystem-wide performance for arbitrary sellers, our computational pipeline isolates the supply chain iteratively. Specifically, for each publisher $s \in \mathcal{S}_{\text{eval}}$ within an evaluation cohort $\mathcal{S}_{\text{eval}} \subset \mathcal{S}$, the algorithms are executed by designating $s$ as the sole active origin node within the global topology. The macro-level performance metrics reported in the subsequent sections represent the statistical aggregation of these independent, single-publisher instances, smoothly bridging the theoretical single-source formulation with broad empirical behavior.

\subsection{Solver Implementation and Operational Architecture}\label{sec:solver_implementation}

By applying the Big-M linearization and the SOS2 piecewise approximation, the NP-hard MINLP is reduced to an MILP. This allows the SPO problem to be solved using exact branch-and-cut algorithms implemented in standard optimization frameworks. The mathematical model is computationally realized in Python using the Pyomo optimization environment \cite{bynum2021pyomo} and solved using the open-source COIN-OR Branch and Cut (CBC) solver \cite{forrest2005cbc}.

To manage the computational trade-off between the precision of the logarithmic NSW approximation and the resulting branching complexity, we modulate the SOS2 approximation grid density across $K \in \{5, 10, 20, 30, 40, 50\}$ breakpoints. To define the minimum strictly positive utility bound for the grid ($U_1$), we set the tolerance scalar to $\epsilon = 10^{-4}$ across all experiments (see Section \ref{sec:piecewise_linear}). A higher $K$ increases the precision of the objective function but requires the solver to branch over a larger set of SOS2 continuous variables (($\lambda_{v,k}$, as defined in Equations \ref{eq:sos2}).)

Furthermore, given the high combinatorial complexity of the supply graphs (exceeding $10^{15}$ feasible paths at $T_{\max} = 3800$~ms) combined with the additional binary variables introduced by the SOS2 piecewise approximations, solving these worst-case instances within strict execution windows is very computationally demanding. Because the MILP operates as an offline strategic benchmark rather than a RTB engine, we constrain the CBC solver with a 300-second wall-clock timeout (and a 1\% relative MIP gap tolerance). This 5-minute budget prevents premature truncation on dense topologies while remaining operationally practical for periodic hourly or daily recalculations. If the solver reaches this limit before proving global optimality, we extract the incumbent solution (the best feasible integer solution found), ensuring SLA-compliant routing decisions across all topologies.

Because MILP solvers cannot meet the sub-150-millisecond limits of real-time OpenRTB auctions, we propose a two-speed operational architecture for production environments:
\begin{enumerate}
    \item \textbf{The Online Tactical Heuristic (K-Shortest Paths):} For real-time per-impression routing, the system uses a K-Shortest Paths (KSP) heuristic based on Yen's algorithm \cite{yen1971finding}. Operating on the pre-processed DAG, for each reachable buyer $b \in B$, the algorithm extracts up to $N=50$ simple paths ordered by ascending cumulative latency. It then discards any path violating the end-to-end $T_{\max}$ timeout or strict participant profitability ($u_v > 0$ for all active nodes), and selects the surviving path across all buyers that maximizes the NSW (marking the run infeasible if no path survives).\looseness=-1
    \item \textbf{The Offline Strategic Benchmark (MILP Solver):} Executed asynchronously outside the real-time auction window (e.g., via hourly or daily batch processing), this exact model evaluates aggregated empirical data to find the mathematically optimal supply paths. While too computationally intensive for live bidding, it provides a rigorous ground-truth baseline to measure the efficiency gap of the online heuristic and enables publishers to strategically prune underperforming intermediaries from their supply chain.
\end{enumerate}

Although this dual architecture defines how the algorithms would operate in a live ad-tech stack, it is important to note that all algorithmic evaluations and performance comparisons presented in the subsequent sections are conducted as computational simulations, testing both the MILP and the KSP heuristic strictly on the static, empirically derived evaluation graphs.

\subsection{Benchmarks and Experimental Setup}

For our experiments, we replace the seller-specific timeouts ($T_{\max,s}$) with a uniform SLA limit $T_{\max,s} = T_{\max}$ for all $s \in \mathcal{S}$. We evaluate network dynamics across three thresholds: $T_{\max} \in \{3000, 3500, 3800\}$~ms.

We evaluate the performance of our MILP optimal routing on the empirical DAG ($F$) against two standard baseline behaviors and our proposed operational heuristic. Across all models, a candidate path $P$ from publisher $s$ to buyer $b \in B$ is feasible if and only if it satisfies both the latency SLA ($L(P) = \sum_{e \in P} t_e \le T_{\max}$) and strict financial profitability ($u_v > 0$ for the buyer, every active intermediary on $P$, and the publisher):
\begin{enumerate}
    \item \textbf{PoA (vs. Decentralized Routing, DEC):} We measure the efficiency loss by comparing uncoordinated routing against our centralized NSW optimum. This baseline represents the local, single-node optimization strategies standard in the literature \cite{balseiro2014, wang2017}. We model this as a myopic greedy heuristic: starting at $s$, each node routes the transaction directly to an adjacent buyer with the highest clearing budget ($\beta_b$) if one is reachable, or otherwise to the immediate intermediary successor with the highest out-degree, completely ignoring cumulative downstream fees and end-to-end latency SLA during path construction. If the greedy walk reaches a dead end or produces a path that violates $L(P) \le T_{\max}$ or $u_v > 0$, the run is marked infeasible.
    \item \textbf{Path Efficiency (vs. Shortest Path Routing, SP):} We compare our fee-aware routing against a standard minimum-hop baseline (SP), which reflects the common industry practice of minimizing hop count under the assumption that shorter chains are more efficient. Specifically, for each reachable buyer $b \in \mathcal{B}$, the algorithm computes the minimum-hop directed path from $s$ to $b$ and selects the single path with the fewest hops across all buyers, breaking ties in favor of the buyer with the highest budget ($\beta_b$). Because this baseline ignores intermediary take-rates ($\gamma$), it often routes traffic through expensive hubs; if no buyer is reachable or the selected path violates $L(P) \le T_{\max}$ or $u_v > 0$, the run is marked infeasible.
    \item \textbf{Operational Heuristic (vs. K-Shortest Paths, KSP):} We compare the exact MILP Oracle against our real-time KSP heuristic introduced in Section~\ref{sec:solver_implementation}, to measure how much of the optimal NSW the heuristic recovers within strict online SLA limits and at what computational speedup.
\end{enumerate}

To quantitatively compare the MILP optimal routing against the isolated baselines, we evaluate each extracted supply path using eight standard performance metrics:
\begin{enumerate}
    \item \textbf{Nash Social Welfare $NSW$:} The global objective value representing the equitable distribution of utility among active participants, defined as:
    \begin{equation}
        NSW = \sum_{v \in \mathcal{V}} w_v
    \end{equation}
    
    \item \textbf{Supply Path Efficiency (SPE):} The proportion of the buyer's actual expenditure that successfully reaches the publisher. It is defined as the ratio between the incoming revenue at the origin publisher node ($s$) and the outgoing spend from the terminal buyer node ($b$). However, as the solution is a single routing path $P$ from $s$ to $b$, this ratio simplifies exactly to the cumulative product of the retention rates across all intermediary nodes on that selected path:
    \begin{equation}
        SPE = \frac{\sum_{e \in OUT(s)} z_e p_e}{\sum_{e \in IN(b)} z_e p_e} = \prod_{v \in P \setminus \{s, b\}} (1 - \gamma_v)
    \end{equation}
    
    \item \textbf{Total Tech Tax ($\tau$):} The cumulative value extracted by intermediary transit nodes (structural friction), calculated as the sum of their utilities relative to the buyer's spend:
    \begin{equation}
        \tau = \frac{\sum_{i \in \mathcal{I}} u_i}{\sum_{e \in IN(b)} z_e p_e} = 1 - SPE
    \end{equation}
    
    \item \textbf{End-to-End Latency ($L$):} The cumulative technical delay of the routed path, verifying technical SLA compliance, measured in milliseconds:
    \begin{equation}
        L = \sum_{e \in \mathcal{A}} z_e t_e
    \end{equation}
    
    \item \textbf{Path Length ($H$):} The total number of active routing edges (hops) in the transaction chain:
    \begin{equation}
        H = \sum_{e \in \mathcal{A}} z_e
    \end{equation}

    \item \textbf{Computational Execution Time ($T_{exec}$):} The wall-clock time required by the MILP solver to terminate within the configured optimality tolerance and time limit, measured in seconds. This metric evaluates how the execution time scales when increasing the SOS2 approximation precision ($K$) across progressively larger network topologies ($T_{\max}$).
    
    \item \textbf{Price of Anarchy ($PoA$):} The quantitative measure of system efficiency lost due to uncoordinated, myopic multi-tier routing. Following applied network routing literature \cite{koutsoupias1999worst, roughgarden2002bad}, we use $PoA$ in an empirical operational sense to measure the coordination gap relative to myopic decentralized routing rather than a formal game-theoretic Nash equilibrium. Because our normalized objective \eqref{eq:objective} computes the sum of logarithms of fractional utilities, the resulting scores are strictly negative. To maintain standard economic interpretation ($PoA \ge 1$), we exponentiate these logarithmic scores to compute the ratio of the actual geometric mean utilities between the centralized optimal routing and the decentralized baseline:
    \begin{equation}\label{eq:poa}
        PoA = \frac{\exp(NSW_{MILP})}{\exp(NSW_{DEC})}
    \end{equation}
    where $NSW_{MILP}$ is the maximum social utility achieved by the exact solver, and $NSW_{DEC}$ is the utility resulting from the DEC heuristic.

    \item \textbf{Feasibility Rate ($FR$):} The percentage of test instances in the evaluation cohort $\mathcal{S}_{\text{eval}}\subset \mathcal{S}$ for which a routing mechanism successfully finds a valid supply path connecting the publisher to an eligible buyer within both the end-to-end latency budget ($L(P) \le T_{\max}$) and strict profitability bounds ($u_v > 0$ for all active nodes):
    \begin{equation}
        FR = \frac{n}{|\mathcal{S}_{\text{eval}}|} \times 100\%
    \end{equation}
    where $n \le |\mathcal{S}_{\text{eval}}|$ denotes the number of feasible instances. Because $PoA$ is defined only when the decentralized baseline yields a valid path, reporting $FR$ alongside $PoA$ captures coordination failures caused by SLA violations. Whenever an algorithm fails to return a feasible path, the run is marked infeasible (contributing $0$ to $FR$ and excluded from path-level metric averages) and all continuous metrics ($NSW, SPE, \tau, L, H$) are recorded as undefined (\texttt{NaN}) rather than zero, preventing artificial distortion of logarithmic utilities ($NSW \le 0$) or path latency and hop averages.
\end{enumerate}

For the experimental scenarios, the intermediary take-rates ($\gamma \in (0, 1)$) are instantiated under the two fee regimes introduced in Section~\ref{sec:data_acquisition}. Under the homogeneous setting, a flat fee of $\gamma = 0.10$ is applied across all intermediaries. Under the heterogeneous setting, each transit node is assigned an independent take-rate drawn from a uniform distribution, $\gamma \sim U(0.05, 0.25)$. This variance forces the algorithms to perform financial arbitrage, demonstrating that the structurally shortest path (SP) does not inherently guarantee the most cost-efficient route.

To stress-test the computational scalability of the MILP formulation across network densities while maintaining identical instance coverage, we construct the evaluation cohort $\mathcal{S}_{\text{eval}} \subset \mathcal{S}$ by extracting the top $|\mathcal{S}_{\text{eval}}| = 50$ highly connected publishers (origin nodes) ranked by out-degree in the baseline graph $F_{3000}$, and evaluate this exact same cohort across $F_{3000}$, $F_{3500}$, and $F_{3800}$. This cohort size ($|\mathcal{S}_{\text{eval}}| = 50$) is independent of the $N = 50$ candidate-path budget per buyer used in the KSP heuristic. Selecting top-degree origin nodes isolates the most computationally demanding sub-topologies, ensuring the solvers are evaluated against the most severe combinatorial bottlenecks. Feasibility rate ($FR$) is computed as a percentage across all $|\mathcal{S}_{\text{eval}}| = 50$ instances, while execution time ($T_{\text{exec}}$) is reported as $\text{Mean} \pm \text{SD}$ over the full cohort. Path-level metrics ($NSW, SPE, \tau, L, H$) are reported as $\text{Mean} \pm \text{SD}$ strictly over each algorithm's $n$ feasible runs. Because logarithmic welfare distributions exhibit heavy lower tails, we additionally report the median and interquartile range (IQR) for $NSW$, and summarize pairwise $PoA$ using the median and IQR over the intersection of instances feasible for both compared algorithms.

We structure our evaluation across three experimental dimensions:
\begin{enumerate}
    \item \textbf{Operational Feasibility and Scalability:} We run the model on a family of network topologies pruned at different SLA constraints ($T_{\max} \in \{3000, 3500, 3800\}$~ms). This allows us to evaluate how financial utility degrades under progressively stricter technical bounds, and we compare the exact MILP execution time against the KSP online heuristic on each topology.
    
    \item \textbf{Comparative Statics and Fee Thresholds:} We perform a dynamic threshold analysis exclusively on the most relaxed topology ($T_{\max} = 3800$~ms). By relaxing the latency constraint, we provide the solver with maximum structural freedom to explore alternative routes. We isolate a central intermediary hub and incrementally increase its take-rate $\gamma$ from a baseline of $\gamma_0 = 0.10$ (while holding the fee of all other nodes constant at this baseline), identifying the exact point where the centralized model redirects traffic to alternative paths to bypass the higher fee.
    
    \item \textbf{Sensitivity Analysis on Synthetic Priors:} Using the same maximally relaxed topology ($T_{\max} = 3800$~ms), we ensure our findings are not reliant on the semi-synthetic estimation of publisher reserves. We perturb the $\sigma$ generation bounds across different percentiles and observe the stability of the core routing decisions.
\end{enumerate}

To enable a fair mathematical comparison of the NSW across all models, we must apply a standardized pricing fallback to the heuristic outputs. While the exact MILP solver dynamically calculates the optimal continuous transaction prices ($p_e$) to maximize system welfare, the baseline algorithms (SP, DEC, and KSP) are strictly structural routing heuristics that do not compute prices natively. 

Because the logarithmic NSW objective acts as an interior-point barrier, it requires strictly positive utility for all active participants ($u_v > 0$). Consequently, a transaction cannot completely exhaust the buyer's maximum valuation ($p_b < \beta_b$). To satisfy this constraint while mirroring real-world RTB dynamics, we parameterize the heuristic clearing price utilizing a standard bid shading assumption \cite{wang2017}. Specifically, we assume the selected buyer retains a 10\% surplus margin ($u_b = 0.10 \beta_b$), injecting 90\% of their maximum budget into the supply path. This continuous price is then multiplicatively degraded by the intermediary take-rates ($\gamma_v$) until it reaches the publisher. While this 90\% allocation anchors the absolute NSW scores for the heuristics, the structural metrics (such as SPE) remain mathematically independent of this assumption, as the initial clearing price cancels out in the efficiency ratios.

\subsection{Results and Discussion}

The computational experiments validate the proposed two-speed operational architecture across the evaluation cohort ($|\mathcal{S}_{\text{eval}}| = 50$), comparing the exact MILP benchmark, the KSP tactical heuristic, the SP structural baseline, and the myopic DEC baseline. Because infeasible transactions yield undefined ($\text{NaN}$) utility and path metrics, sample means across all figures and tables (NSW, SPE, Tech Tax, path depth, and latency) are computed strictly over each algorithm's feasible subset ($n = 48$--$49/50$ for MILP and KSP, $n = 46/50$ for SP, and $n = 12$--$13/50$ for DEC). Consequently, unaligned comparisons are subject to systematic survivorship bias: algorithms with lower feasibility rates (SP and especially DEC) exclude difficult, low-budget publishers with multi-hop supply chains, artificially inflating their apparent averages on easier instances. To eliminate this selection effect, all pairwise welfare claims are verified strictly over common feasible intersections using paired Wilcoxon signed-rank tests.

\subsubsection{Computational Scalability and Heuristic Validation}
To evaluate the computational overhead of the SOS2 approximation of the logarithmic objective and validate the online tactical heuristic, we measure execution time ($T_{\text{exec}}$) across the evaluation cohort $|\mathcal{S}_{\text{eval}}| = 50$. First, we run the exact MILP solver across all grid densities $K \in \{5, 10, 20, 30, 40, 50\}$ under varying latency constraints, configuring the solver with a 300-second timeout and a 1\% relative optimality gap. Second, we evaluate the execution time of the KSP heuristic to verify whether it can deliver near-optimal NSW within real-time programmatic bidding windows.

\begin{figure}[htbp]
    \centering
    \includegraphics[width=0.68\textwidth]{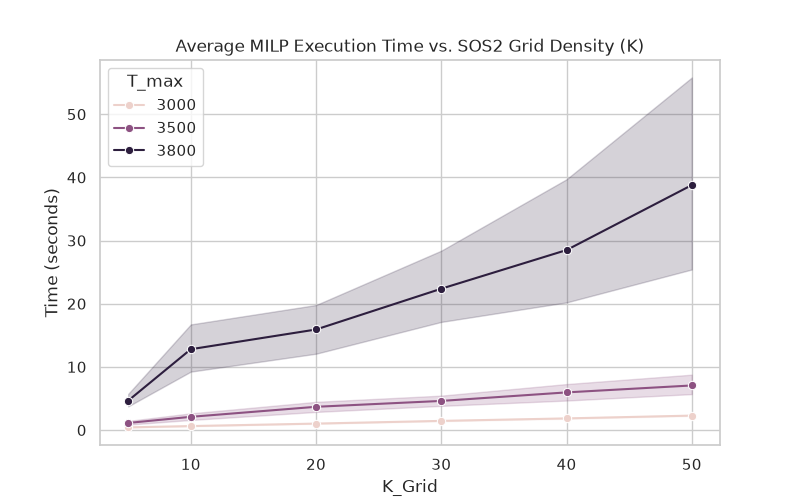}
    \caption{Average MILP Execution Time vs. SOS2 Grid Density ($K$) across the evaluation cohort ($|\mathcal{S}_{\text{eval}}| = 50$ publishers). Shaded bands denote 95\% confidence intervals.}
    \label{fig:exec_time}
\end{figure}

As shown in Figure~\ref{fig:exec_time}, MILP execution time scales with both the SOS2 grid density ($K$) and the topological relaxation ($T_{\max}$). On the most constrained network ($T_{\max} = 3000$~ms, 754 directed arcs), the solver resolves instances in approximately 1.02~seconds at the baseline density $K=20$ and around 2.4~seconds at $K=50$. On the maximally relaxed topology ($T_{\max} = 3800$~ms, 2,854 arcs), finer logarithmic discretization activates more competing SOS2 segments in the branch-and-bound tree, increasing average execution time from 15.92~seconds at $K=20$ to approximately 38.8~seconds at $K=50$. Accordingly, $K=20$ represents an effective operational trade-off between approximation fidelity and offline computation time.

While the MILP solver serves as an offline strategic benchmark, the KSP heuristic is designed for sub-second execution. Operating on the pre-processed DAG topologies, KSP recorded an average execution time of 2.4~ms at $T_{\max} = 3000$~ms, 159.6~ms at $T_{\max} = 3500$~ms, and 363.6~ms at $T_{\max} = 3800$~ms. These results confirm the heuristic's practical viability for real-time programmatic path scoring.

\subsubsection{Objective Optimization and Financial Distribution}
\begin{figure}[htbp]
    \centering
    \includegraphics[width=0.68\textwidth]{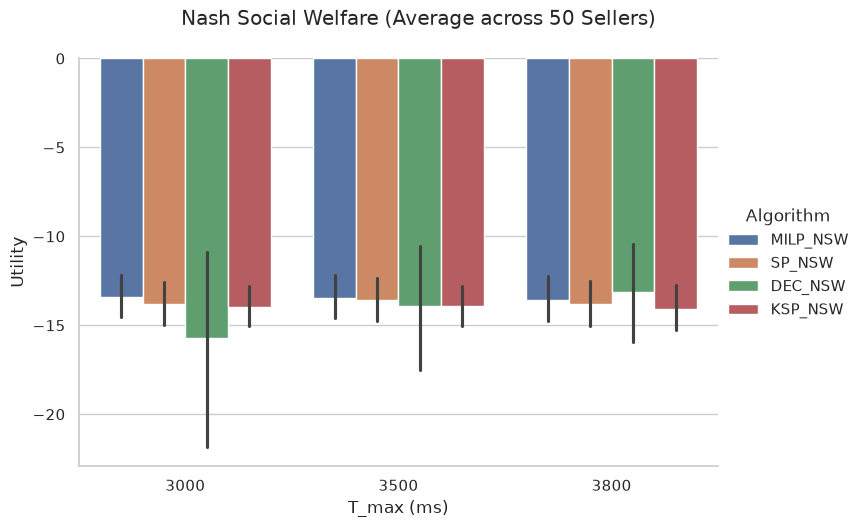}
    \caption{NSW comparison across routing algorithms for different $T_{\max}$ thresholds, averaged over feasible instances in $|\mathcal{S}_{\text{eval}}| = 50$ (with KSP path budget $N = 50$ and MILP grid density $K = 20$). Error bars denote 95\% confidence intervals. Note that DEC averages reflect only the small subset of high-margin feasible instances ($n = 12$--$13/50$), whereas MILP and KSP cover $n = 48$--$49/50$.}
    \label{fig:nsw_comp}
\end{figure}

\begin{figure}[htbp]
    \centering
    \includegraphics[width=0.68\textwidth]{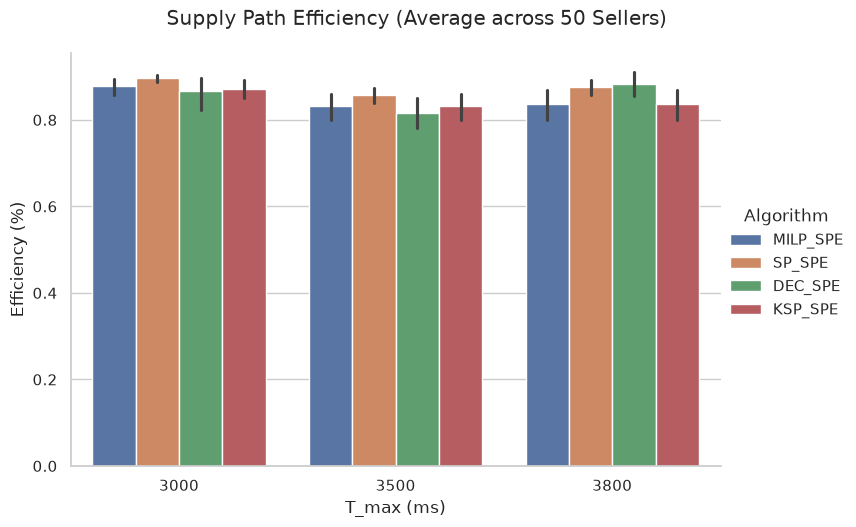}
    \caption{Supply Path Efficiency (SPE) across routing algorithms ($K=20$, $N=50$), averaged over each method's feasible instances ($n = 48$--$49$ for MILP/KSP, $46$ for SP, $12$--$13$ for DEC). Error bars denote 95\% confidence intervals.}
    \label{fig:spe_comp}
\end{figure}

\begin{figure}[htbp]
    \centering
    \includegraphics[width=0.68\textwidth]{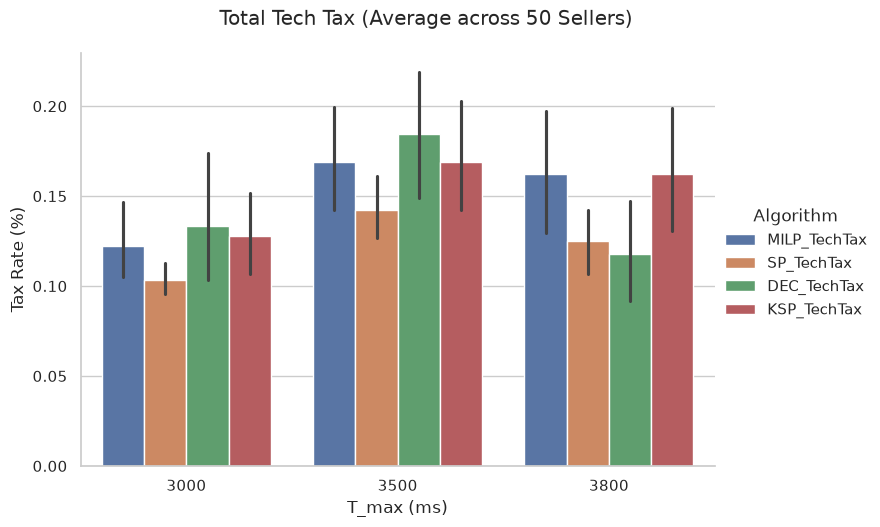}
    \caption{Total Tech Tax ($\tau = 1 - \text{SPE}$) across routing algorithms ($K=20$, $N=50$), averaged over each method's feasible instances ($n = 48$--$49$ for MILP/KSP, $46$ for SP, $12$--$13$ for DEC). Error bars denote 95\% confidence intervals.}
    \label{fig:techtax_comp}
\end{figure}

To evaluate how each mechanism navigates asymmetric intermediary take-rates ($\gamma \sim U(0.05, 0.25)$), we examine aggregate welfare and fee extraction at the baseline grid density $K=20$ (summarized in Table~\ref{tab:main_results} and illustrated in Figures~\ref{fig:nsw_comp}--\ref{fig:techtax_comp}).

\begin{table}[htbp]
\centering
\caption{Aggregate Performance Metrics Across Routing Mechanisms ($K=20$, Heterogeneous Fees $\gamma \sim U(0.05, 0.25)$). Continuous metrics (NSW, SPE, Tech Tax, Latency, and Hops) are reported as Mean $\pm$ SD over feasible runs. For NSW, the median and IQR are additionally provided.}\label{tab:main_results}
\tiny
\setlength{\tabcolsep}{1.5pt}
\begin{tabular*}{\textwidth}{@{\extracolsep{\fill}}llccccccc}
\hline
\textbf{$T_{\max}$} & \textbf{Method} & \textbf{FR} & \textbf{NSW} & \textbf{NSW} & \textbf{SPE} & \textbf{Tech Tax} & \textbf{Latency} & \textbf{Hops} \\
\textbf{(ms)} & & \textbf{(\%)} & & \textbf{Med. (IQR)} & & \textbf{$\tau$} & \textbf{(ms)} & \\ \hline
\multirow{4}{*}{3000} & \textbf{MILP} & 96.0 & $\mathbf{-13.42 \pm 4.23}$ & $\mathbf{-12.57}$ (6.40) & $0.878 \pm 0.075$ & $0.122 \pm 0.075$ & $1315 \pm 901$ & $2.13 \pm 0.49$ \\
 & \textbf{KSP} & 96.0 & $-13.95 \pm 4.19$ & $-13.15$ (6.11) & $0.872 \pm 0.082$ & $0.128 \pm 0.082$ & $1307 \pm 890$ & $2.17 \pm 0.56$ \\
 & \textbf{SP} & 92.0 & $-13.83 \pm 4.26$ & $-13.15$ (2.86) & $0.896 \pm 0.029$ & $0.104 \pm 0.029$ & $1207 \pm 879$ & $2.00 \pm 0.00$ \\
 & \textbf{DEC} & 26.0 & $-15.69 \pm 10.61$ & $-13.16$ (6.85) & $0.867 \pm 0.071$ & $0.133 \pm 0.071$ & $1196 \pm 611$ & $2.15 \pm 0.55$ \\ \hline
\multirow{4}{*}{3500} & \textbf{MILP} & 98.0 & $\mathbf{-13.44 \pm 4.46}$ & $\mathbf{-13.08}$ (5.88) & $0.831 \pm 0.108$ & $0.169 \pm 0.108$ & $1352 \pm 930$ & $2.18 \pm 0.63$ \\
 & \textbf{KSP} & 98.0 & $-13.93 \pm 4.40$ & $-13.55$ (5.76) & $0.831 \pm 0.108$ & $0.169 \pm 0.108$ & $1352 \pm 930$ & $2.18 \pm 0.63$ \\
 & \textbf{SP} & 92.0 & $-13.60 \pm 4.31$ & $-13.54$ (2.84) & $0.857 \pm 0.059$ & $0.143 \pm 0.059$ & $1207 \pm 879$ & $2.00 \pm 0.00$ \\
 & \textbf{DEC} & 26.0 & $-13.93 \pm 6.66$ & $-13.55$ (6.85) & $0.816 \pm 0.069$ & $0.184 \pm 0.069$ & $1049 \pm 483$ & $2.00 \pm 0.00$ \\ \hline
\multirow{4}{*}{3800} & \textbf{MILP} & 98.0 & $\mathbf{-13.56 \pm 4.59}$ & $\mathbf{-13.15}$ (7.09) & $0.838 \pm 0.129$ & $0.163 \pm 0.129$ & $1345 \pm 919$ & $2.22 \pm 0.69$ \\
 & \textbf{KSP} & 98.0 & $-14.07 \pm 4.52$ & $-13.75$ (6.70) & $0.838 \pm 0.129$ & $0.163 \pm 0.129$ & $1345 \pm 919$ & $2.22 \pm 0.69$ \\
 & \textbf{SP} & 92.0 & $-13.78 \pm 4.48$ & $-13.75$ (3.41) & $0.875 \pm 0.068$ & $0.125 \pm 0.068$ & $1207 \pm 879$ & $2.00 \pm 0.00$ \\
 & \textbf{DEC} & 24.0 & $-13.12 \pm 5.21$ & $-13.75$ (6.23) & $0.882 \pm 0.050$ & $0.118 \pm 0.050$ & $1086 \pm 484$ & $2.00 \pm 0.00$ \\ \hline
\end{tabular*}
\end{table}

Across all latency regimes, the exact MILP formulation achieves the highest NSW. Consistent with the survivorship bias noted above, the 2--3 low-budget publishers rescued exclusively by KSP yield low residual utilities (mean NSW $\approx -19.1$ to $-20.3$), pulling KSP's unaligned mean slightly below SP's mean over the 46 easier instances, while DEC's apparent unaligned mean at $T_{\max}=3800$~ms ($-13.12$) reflects only the 12 most favorable publishers ($24\%$ feasibility). By contrast, MILP's endogenous price optimization generates sufficient surplus gains to surpass SP even in unaligned averages. Evaluated strictly over common feasible instances via paired Wilcoxon signed-rank tests, MILP significantly outperforms KSP across all topologies ($+0.536$, $p = 7.11 \times 10^{-15}$ at $T_{\max}=3000$~ms; $+0.487$, $p = 3.55 \times 10^{-15}$ at $T_{\max}=3500$~ms; $+0.519$, $p = 3.55 \times 10^{-15}$ at $T_{\max}=3800$~ms), SP ($+0.649$, $+0.601$, and $+0.642$, $p = 2.84 \times 10^{-14}$), and DEC ($+4.624$, $+2.721$, $p = 2.44 \times 10^{-4}$; $+2.251$, $p = 4.88 \times 10^{-4}$). Likewise, on their common feasible subset ($n=46$), KSP dominates or matches SP on 100\% of instances, improving mean NSW by $+0.091$ to $+0.098$ logarithmic units.

A key structural insight emerges by comparing MILP and KSP at $T_{\max} \in \{3500, 3800\}$~ms: both algorithms select identical physical supply paths (yielding identical mean SPE of $0.831$--$0.838$, latency, and hop counts), yet MILP achieves an NSW advantage of approximately $+0.49$ to $+0.52$ logarithmic units. Whereas KSP relies on exogenous equal-surplus clearing prices along candidate routes, the MILP formulation endogenously optimizes clearing prices $p_u$ to balance utility across the publisher, intermediaries, and winning buyer under the concave Nash product. Conversely, while SP achieves slightly higher unaligned SPE ($0.857$--$0.896$) by restricting itself to 2-hop paths, it ignores downstream buyer budgets $\beta_b$, missing higher-value demand accessible through multi-hop routing.

\subsubsection{Network Mechanics (Path Depth and Latency)}
\begin{figure}[htbp]
    \centering
    \includegraphics[width=0.68\textwidth]{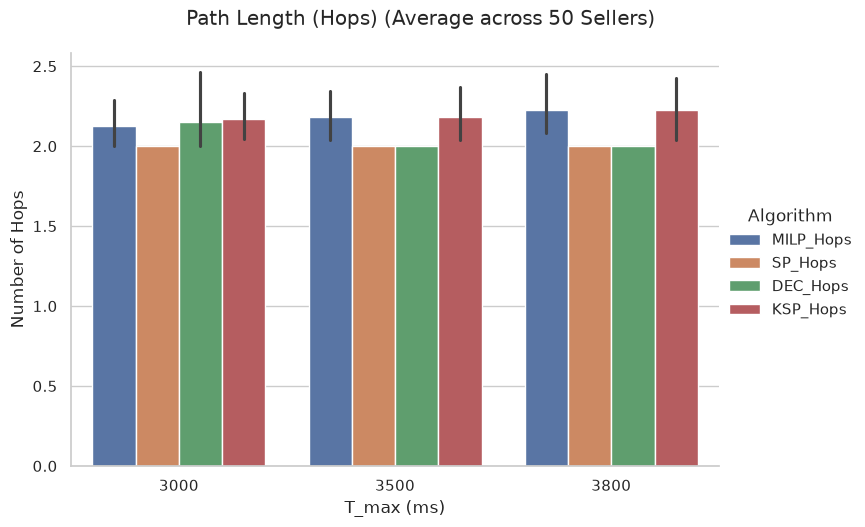}
    \caption{Supply path length ($H$) across routing algorithms ($K=20$, $N=50$), averaged over each method's feasible instances ($n = 48$--$49$ for MILP/KSP, $46$ for SP, $12$--$13$ for DEC). Error bars denote 95\% confidence intervals.}
    \label{fig:hops_comp}
\end{figure}

\begin{figure}[htbp]
    \centering
    \includegraphics[width=0.68\textwidth]{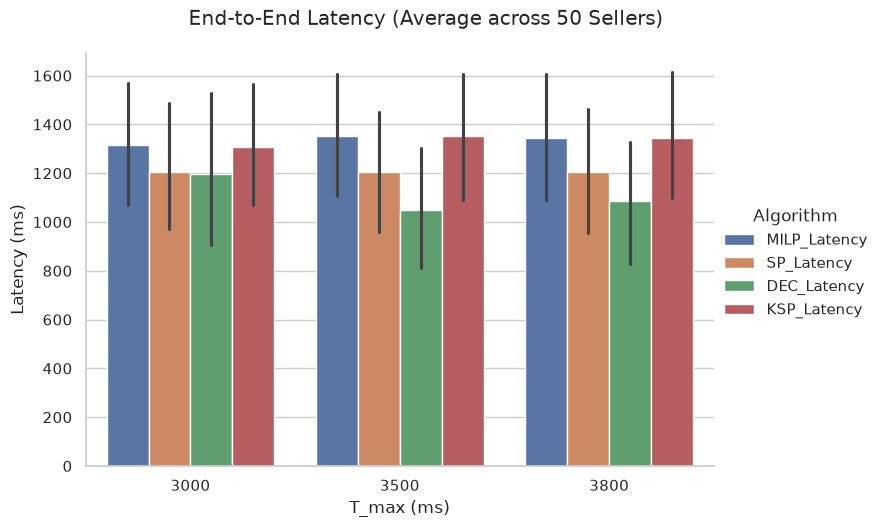}
    \caption{End-to-end latency ($L$) across routing algorithms ($K=20$, $N=50$), averaged over each method's feasible instances ($n = 48$--$49$ for MILP/KSP, $46$ for SP, $12$--$13$ for DEC). Error bars denote 95\% confidence intervals.}
    \label{fig:latency_comp}
\end{figure}

To understand the structural decisions driving these financial outcomes, we analyze the path depth ($H$) and end-to-end latency ($L$) generated by each algorithm (Figures~\ref{fig:hops_comp} and~\ref{fig:latency_comp}).

Under heterogeneous fees, the SP baseline strictly bounds every feasible supply chain to $2.00$ hops, yielding a constant average end-to-end latency of $1206.7$~ms across all $T_{\max}$ thresholds. However, this restriction to minimum-hop paths prevents SP from reaching high-budget buyers connected via secondary exchanges, limiting its feasibility rate to 92\%.

To access higher-paying buyers or avoid high-fee intermediaries, the MILP and KSP models selectively extend the path to an average depth of $2.13$--$2.22$ hops. This controlled topological expansion increases average end-to-end latency modestly to $1307$--$1352$~ms, remaining well within standard OpenRTB latency budgets while increasing feasibility to 96\%--98\% and unlocking substantially higher NSW. Meanwhile, the few transactions where the myopic DEC heuristic survives ($24\%$--$26\%$) are restricted to short, low-latency paths ($2.00$--$2.15$ hops, $1048.5$--$1196.4$~ms), because longer uncoordinated chains fail budget or latency feasibility entirely.

\subsubsection{Systemic Inefficiency and Price of Anarchy}

\begin{figure}[htbp]
    \centering
    \includegraphics[width=0.98\textwidth]{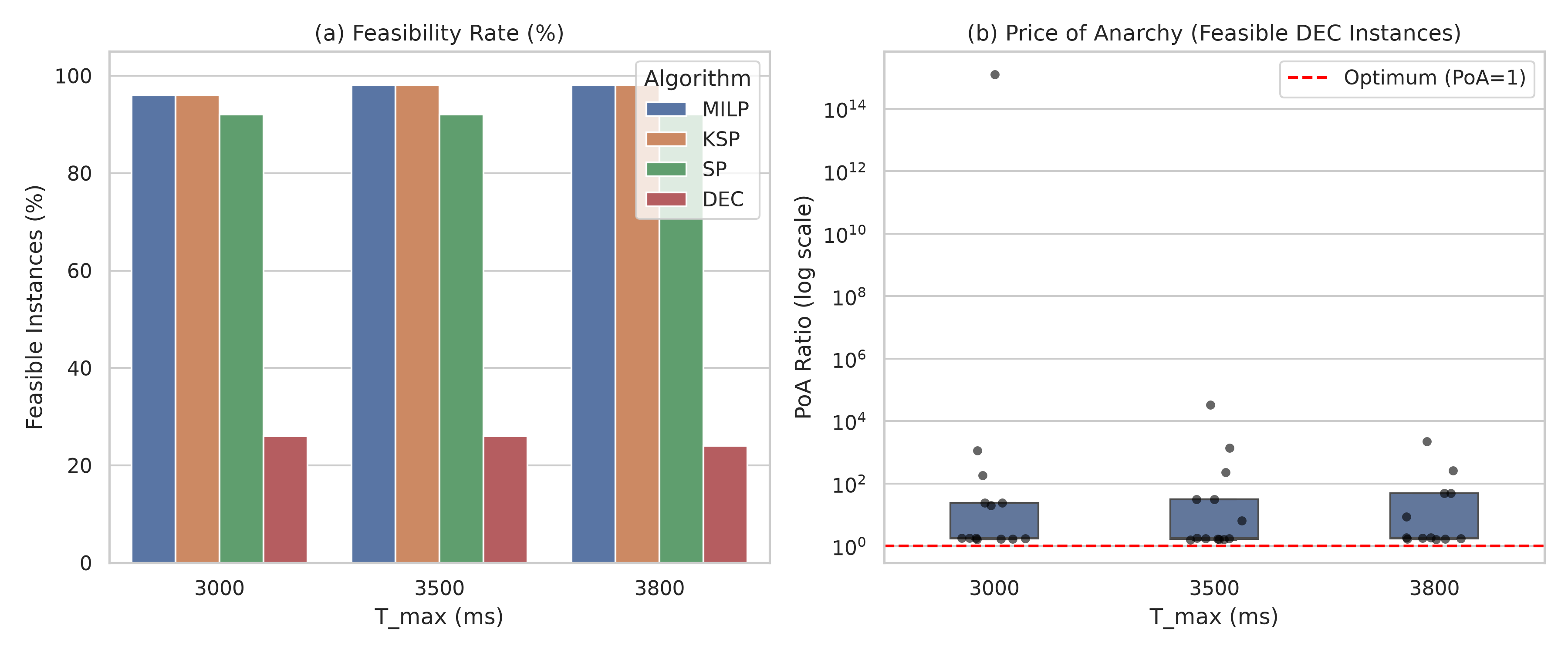}
    \caption{Systemic inefficiency of decentralized routing. (a) Feasibility Rate ($FR$, \%) tracking the proportion of latency-compliant and financially viable paths across the evaluation cohort ($|\mathcal{S}_{\text{eval}}| = 50$ publishers). (b) Log-scaled PoA ($\exp(NSW_{\text{MILP}}) / \exp(NSW_{\text{DEC}})$) distribution exclusively over the $n$ instances where DEC yielded a feasible path; boxplots show the median and IQR, while black markers represent individual feasible publisher instances.}
    \label{fig:poa_comp}
\end{figure}

As illustrated in Figure~\ref{fig:poa_comp}, evaluating the PoA~\eqref{eq:poa} for the decentralized baseline (DEC) highlights two distinct operational failures: widespread transaction infeasibility and severe welfare loss among clearing transactions.

First, the myopic, tier-by-tier decisions of the DEC baseline cause the vast majority of transactions to fail. Figure~\ref{fig:poa_comp}a shows that DEC achieves a feasibility rate of only 26.0\% ($13/50$) at $T_{\max} \in \{3000, 3500\}$~ms and drops to 24.0\% ($12/50$) at $T_{\max} = 3800$~ms, whereas centralized MILP optimization and KSP achieve 96.0\%--98.0\% feasibility ($48$--$49/50$). Without global look-ahead, local greedy forwarding frequently routes into dead-end paths that either exceed $T_{\max}$ or violate downstream reserve prices ($\sigma_s > p_u$) due to uncoordinated fee stacking, resulting in undefined ($NaN$) PoA values for roughly three-quarters of the cohort.

Second, on the small subset of instances where DEC clears ($n = 12$ to $13$), the resulting PoA ratio exhibits heavy-tailed exponential degradation (Figure~\ref{fig:poa_comp}b). The median PoA remains consistently elevated across all topologies, reaching $1.80$ ($\text{IQR} = 22.01$) at $T_{\max} = 3000$~ms, $1.79$ ($\text{IQR} = 29.68$) at $T_{\max} = 3500$~ms, and $1.82$ ($\text{IQR} = 46.72$) at $T_{\max} = 3800$~ms. This indicates that even in typical clearing cases, centralized coordination nearly doubles the geometric mean utility compared to decentralized routing. In extreme tail instances at $T_{\max} = 3000$~ms, cascading margin extraction reduces residual publisher utility to near-zero levels, driving individual PoA ratios above $10^{15}$. These results confirm that uncoordinated multi-tier routing severely degrades both market clearing reliability and economic welfare.

\subsubsection{Comparative Statics and Fee Thresholds}
To evaluate how the network adapts to aggressive intermediary pricing, we conduct a dynamic threshold analysis across 50 valid publisher-hub pairs in $F_{3800}$. These pairs were extracted deterministically by prioritizing publishers with the highest network out-degree, restricted to instances where an alternative feasible bypass path exists upon hub removal. The procedure is executed exclusively on the maximally relaxed topology ($T_{\max} = 3800$~ms) to ensure a sufficient number of feasible bypass routes. First, we run the MILP solver at a uniform 10\% baseline fee to identify the optimal path and its highest-volume intermediary hub. Second, we programmatically remove this hub from the network and resolve the MILP to calculate the SPE of the next-best route. Finally, we restore the hub and incrementally increase its specific take-rate in discrete 10\% steps, holding all other nodes at 10\%. The hub is formally considered ``excluded'' at the exact threshold where the MILP solver redirects traffic to the alternative path.

\begin{table}[htbp]
\centering
\caption{Distribution of Intermediary Hub Exclusion Thresholds ($T_{\max} = 3800$~ms)\label{tab:exclusion}}
{\begin{tabular}{lc}
\hline
\textbf{Fee Threshold ($\gamma$)} & \textbf{Frequency (Number of Hubs Excluded)} \\ \hline
10\% & 9 \\
20\% & 20 \\
30\% & 14 \\
40\% & 7 \\
50\% & 0 \\
60\% & 0 \\ \hline
\multicolumn{2}{l}{\textbf{Average Dynamic Exclusion Threshold:} \textbf{23.8\%}} \\ \hline
\end{tabular}}
\end{table}

The threshold analysis reveals a highly sensitive routing equilibrium (Table~\ref{tab:exclusion}). When all nodes operate at a baseline 10\% take-rate, the MILP identifies structurally efficient paths. However, as the central hub attempts to extract higher margins, the exact solver rapidly redirects flow. Across the 50 evaluated pairs, 58\% of the dominant intermediaries (29 out of 50) were excluded from the optimal supply path the moment their take-rate reached 20\% or lower. The global average exclusion threshold across the 50 pairs was 23.8\%, with no central hub retaining traffic at a 40\% take-rate. This provides empirical confirmation that programmatic ad-networks possess sufficient structural redundancy to bypass excessive fees, provided buyers route dynamically rather than statically locking into singular DSP-SSP partnerships.

By isolating a single hub and holding the rest of the network constant (\emph{ceteris paribus}), this threshold analysis strictly measures the structural tolerance of the topology, identifying the maximum pricing power a node possesses before redundancy forces its exclusion, independent of competitor price cuts.

\subsubsection{Sensitivity Analysis on Synthetic Priors}
To verify the robustness of our findings against variations in the semi-synthetic publisher reserve parameter ($\sigma_s$), we conduct a two-tiered sensitivity analysis on the relaxed baseline topology ($T_{\max} = 3800$~ms). We perturb $\sigma_s$ from 25\% to 200\% of its original estimated value. At the macro level, we evaluate the entire cohort of 50 instances to determine if the network structurally redirects traffic to cheaper paths when the reserve price rises. At the micro level, we trace a single representative transaction to observe the exact internal mechanism the MILP solver uses to balance the system when the topology is held constant.

At the macro level, the routing paths proved highly resilient. Aggregating the solver results across all 400 perturbed scenarios (50 seller-buyer instances evaluated across 8 discrete multiplier steps), the optimal physical paths remained perfectly stable (100.0\%) regardless of the reserve-price multiplier. The network completely absorbed the simulated financial volatility without forcing any structural fragmentation or rerouting across the entire evaluation cohort.

\begin{figure}[htbp]
    \centering
    \includegraphics[width=0.75\textwidth]{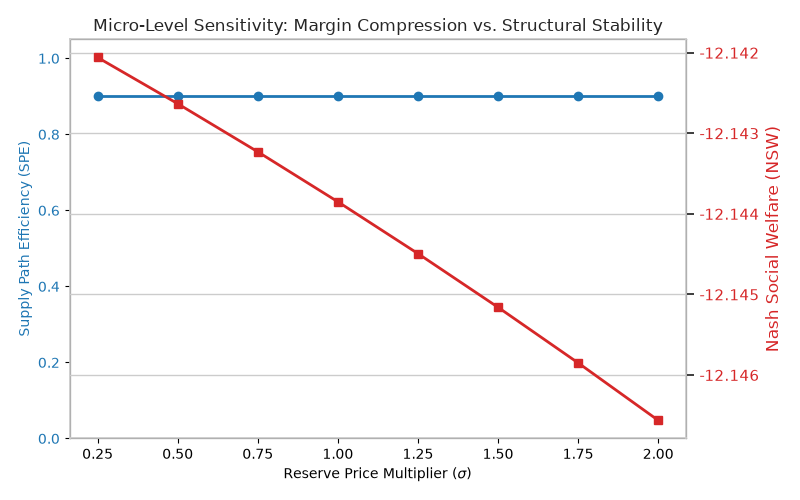}
    \caption{Micro-level sensitivity for \textit{detik.com} ($T_{\max} = 3800$~ms): margin compression in NSW versus structural stability in SPE across reserve-price multipliers from 25\% to 200\%.}
    \label{fig:sensitivity_compression}
\end{figure}

At the micro level, we analyze the specific interaction between the initiating seller (\textit{detik.com}) and its winning buyer (\textit{teads}). This specific pair was deterministically selected for detailed tracing because it represents the publisher with the highest network out-degree that yields a feasible baseline solution, ensuring our micro-level observations are drawn from the most highly connected routing topology in the cohort. The exact mechanism of stabilization is illustrated in Figure~\ref{fig:sensitivity_compression}. As the reserve price $\sigma_s$ is incrementally perturbed from 25\% ($0.0135$~CPM) to 200\% ($0.1076$~CPM) of its baseline value, the physical routing efficiency (SPE) remains rigidly flat at $0.900$. Instead of redirecting traffic to suboptimal paths, the solver accommodates the rising publisher floor by smoothly compressing margins within the continuous clearing price variables ($p_u$). This financial compression is precisely captured by the optimal NSW, which adjusts monotonically from $-12.142$ to $-12.147$. 

Together, these results demonstrate that the network's structural equilibrium is mathematically independent of the synthetic boundaries used during evaluation. Consequently, our prior conclusions regarding intermediary exclusion thresholds are driven entirely by the topology itself, rather than the estimated reserve parameters.\looseness=-1

\subsection{Managerial Insights and Policy Implications}

The results show that the MILP approach establishes a highly efficient functional balance across the programmatic ecosystem. While the SP baseline minimizes topological distance at exactly two hops, its financial blindness restricts it to a 92.0\% feasibility rate and prevents it from accessing higher-value demand connected through secondary exchanges, reducing overall NSW. Conversely, the myopic DEC baseline suffers widespread market failure across all latency regimes (74.0\%--76.0\% infeasibility), collapsing under uncoordinated cascading fees and latency overruns. The MILP model provides a practical strategic benchmark, proving that global optimization is computationally viable: it maximizes joint economic welfare (NSW) through endogenous price clearing while guaranteeing that transactions clear within strict technical limits, while the KSP heuristic operationalizes these path selections within sub-second bidding windows.

Based on these findings, we provide concrete guidelines for platforms, publishers, and regulators:
\begin{enumerate}
    \item \textbf{Fee Capping and Threshold Management:} Platform managers can use this optimization model to pinpoint when intermediary fees become structural roadblocks. Our threshold analysis indicates that when a central intermediary's take-rate rises excessively (averaging 23.8\% across evaluated hubs, with 58\% excluded at or below 20\%), the routing logic actively bypasses that node to preserve system welfare. Platforms should implement competitive fee caps to prevent routing exclusion.
    \item \textbf{Supply Chain Trimming ($k=4$ Limit):} Empirical testing confirms that optimal transaction paths require at most 4 hops (with realized optimal paths averaging 2.13--2.22 hops). Publishers and buyers should configure their systems to automatically drop transaction chains exceeding this depth to prevent SLA timeouts without sacrificing financial utility.
    \item \textbf{Pricing Shocks and Route Stability:} Our macro-micro sensitivity analysis reveals that optimal physical supply paths are 100.0\% robust across the evaluation cohort to substantial perturbations in publisher reserve prices ($\sigma_s$ from 25\% to 200\% of baseline). When publishers adjust reserve floors, the network absorbs the shock through internal margin compression in clearing prices rather than structural rerouting. Network managers do not need to constantly reconfigure physical SSP/DSP connections in response to routine reserve-price volatility.
    \item \textbf{Policy Mandates for Fee Transparency:} Current standards such as \texttt{ads.txt} and \texttt{sellers.json} provide structural path transparency but omit contractual take-rates. Our simulations demonstrate that this fee opacity drives severe market failure and high Price of Anarchy in decentralized routing. To enable self-correction toward efficient equilibria, industry consortiums should mandate standardized take-rate disclosures within OpenRTB payloads.
\end{enumerate}

\section{Conclusion}\label{sec:conclusion}

The opacity of the programmatic advertising ecosystem has historically restricted SPO to fragmented heuristics. We challenged these myopic models by formalizing the multi-tier supply chain as a DAG and proposing a two-speed operational architecture centered on a MILP formulation with logarithmic SOS2 approximation to maximize system-wide NSW, paired with our proposed KSP tactical heuristic for real-time execution.

Our empirical evaluations on OpenRTB datasets demonstrate that uncoordinated routing suffers from severe market failure and a heavy-tailed Price of Anarchy. Across all evaluated latency regimes ($T_{\max} \in \{3000, 3500, 3800\}$~ms), myopic decentralized algorithms fail in 74\%--76\% of transactions due to cascading fee stacking and latency overruns, while nearly halving typical geometric welfare even when transactions clear. Furthermore, while Shortest Path heuristics minimize hop depth, their inability to account for downstream buyer budgets caps feasibility at 92\% and forfeits substantial surplus compared to multi-hop routing. By contrast, our MILP formulation computes an exact equilibrium that endogenously balances clearing prices, ensuring strict SLA compliance, 96\%--98\% feasibility, and statistically significant NSW gains ($p < 10^{-13}$), while KSP closely approximates these path selections within real-time bidding windows. Moreover, sensitivity analysis confirms 100.0\% structural routing stability across the evaluation cohort against substantial reserve-price perturbations, demonstrating that optimal topologies absorb financial volatility through internal margin compression rather than path fragmentation.

For practitioners, preventing market failure requires capping central intermediary take-rates (where average exclusion occurs at 23.8\%) and bounding supply chains to a maximum of 4 hops. A limitation of our experimental design in deriving this threshold is that the fee rate $\gamma$ is set exogenously for a single central hub, holding the rest of the network's pricing structure constant (\emph{ceteris paribus}). By isolating a single hub, our approach measures the structural tolerance of the routing topology, identifying the maximum pricing power a central node possesses before network redundancy forces its exclusion, independent of competitor price cuts. However, in real-world ad-tech markets, the determination of $\gamma$ is endogenous and strategic, functioning as a non-cooperative pricing game. While our current model abstracts away these competitive dynamics---alongside operational bid throttling and privacy constraints---to maintain tractability, future research should transition the static threshold analysis into a dynamic game-theoretic model to find Nash equilibria within latency constraints. Additionally, expanding the framework to include capacitated network flows and distributed optimization techniques (such as Lagrangian relaxation) will enable privacy-preserving coordination under asymmetric cost structures.


\bibliographystyle{plain}
\bibliography{sn-bibliography}

\end{document}